# Predicting Subsurface Abnormalities Growth using Physics-Informed Neural Networks

**Mehrdad S. Dizaji, Ph.D.**
Post-Doctoral Research Fellow, Federal Highway Administration,
Turner-Fairbank Highway Research Center, McLean, VA 22101,
m.shafiei.dizaji.ctr@dot.gov

**Hoda Azari, Ph.D.**
Nondestructive Evaluation Program and Laboratory Manager,
Federal Highway Administration,
Turner-Fairbank Highway Research Center, McLean, VA 22101,
hoda.azari@dot.gov

**Abstract**
The research explores the pioneering integration of Physics-Informed Neural Networks (PINNs) into the domain of Ground-Penetrating Radar (GPR) data prediction, akin to advancements in medical imaging for tracking tumor progression in the human body. This research presents a detailed development framework for a specialized PINN model, proficient at interpreting and forecasting GPR data, much like how medical imaging models predict tumor behavior. By harnessing the synergy between deep learning algorithms and the physical laws governing subsurface structures—or in medical terms, human tissues—the model effectively embeds the physics of electromagnetic wave propagation into its architecture. This ensures that predictions not only align with fundamental physical principles but also mirror the precision needed in medical diagnostics for detecting and monitoring tumors. The suggested deep learning structure comprises three components: a CNN, a spatial feature channel attention (SFCA) mechanism, and ConvLSTM, along with temporal feature frame attention (TFFA) modules. The attention mechanism computes channel attention and temporal attention weights using self-adaptation, thereby fine-tuning the visual and temporal feature responses to extract the most pertinent and significant visual and temporal features. By integrating physics directly into the neural network, our model has shown enhanced accuracy in forecasting GPR data. This improvement is vital for conducting effective assessments of bridge deck conditions and other evaluations related to civil infrastructure. The use of Physics-Informed Neural Networks (PINNs) has demonstrated the potential to transform the field of Non-Destructive Evaluation (NDE) by enhancing the precision of infrastructure deterioration predictions. Moreover, it offers a deeper insight into the fundamental mechanisms of deterioration, viewed through the prism of physics-based models.

***Keywords:*** Physics-Informed Neural Networks, Deep Learning, Ground-Penetrating Radar (GPR), NDE, ConvLSTM, Physics, Data driven

## 1. Introduction

Bridge decks serve as the primary defense against traffic loads and environmental factors, deteriorating more rapidly than other parts of a bridge. Consequently, they consume the largest share of resources allocated for maintenance, repair, and replacement (MRR) of bridge infrastructure. Current methods of assessing their condition and modeling deterioration primarily depend on visual inspection (VI) results and subsequent condition ratings. However, research has exposed several drawbacks of VI, including subjectivity, reliability, and lack of comprehensiveness [1]. Nondestructive evaluation (NDE) offers detailed and unbiased insights that are fundamentally unattainable through visual inspection alone. It enables accurate quantification and monitoring of bridge deck deterioration. Research indicates that NDE measurements are much more sensitive to internal deterioration compared to condition ratings from the National Bridge Inventory (NBI) [2]. However, current approaches to analyzing NDE surveys often involve evaluating the condition of each point independently, without considering its relationship to adjacent points, other NDE methods, or measurements from different time points. This limitation affects the reliability, interpretability, and ability to forecast deterioration using existing NDE data processing methods. This study aims to enhance the reliability and interpretability of NDE results by incorporating overlooked contextual information, specifically the spatial and temporal context. In essence, while identifying deterioration based on a single isolated data point within a single NDE scan may yield uncertain results, analyzing the patterns and consistency between a data point and its neighboring points in space and time can significantly increase confidence in the insights derived from NDE surveys. Figure 1 schematically depicts the type of context information. The proposed research aims to answer the question:

*"How can we utilize machine (deep) learning techniques to extract insights from spatial and temporal patterns?"*

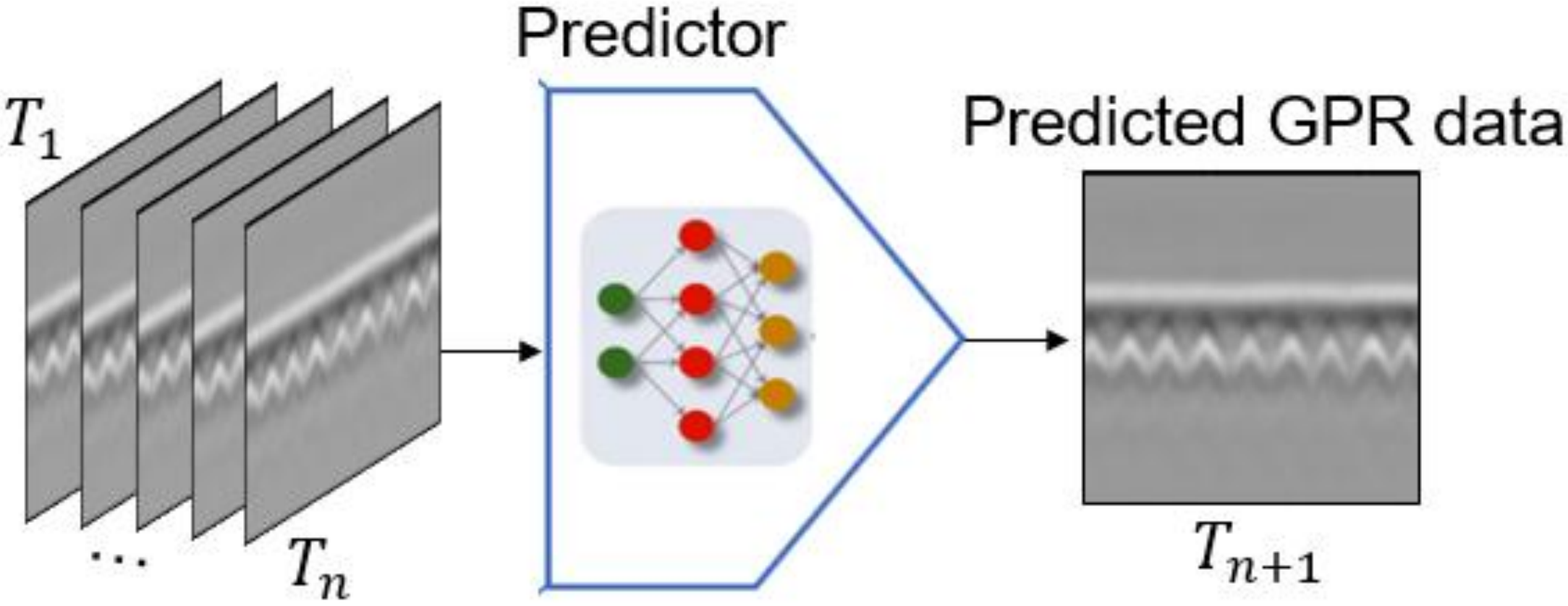

Figure 1. Schematically depicts the type of context information.

*1.1. NDE for Condition Assessment of Infrastructure*

Various technologies have been suggested and examined for the nondestructive evaluation of concrete structures. Some of the most commonly used techniques include impact echo (IE), electrical resistivity (ER), ground-penetrating radar (GPR), half-cell potential (HCP), ultrasonic surface waves (USW), ultrasonic testing (UT), impulse response (IR), and infrared thermography (IRT) [3], [4]. A collaborative study conducted by the Federal Highway Administration (FHWA) investigated and compared a diverse range of nondestructive evaluation (NDE) technologies for assessing the condition of bridge decks, both in real-world field conditions and laboratory settings [4]. Lin et al. conducted research to analyze how various types of overlays influenced the performance of nine nondestructive evaluation (NDE) technologies in a controlled laboratory experiment [3]. Robinson et al. investigated the pairwise correlation coefficients of several nondestructive evaluation (NDE) methods performed on three bridges in Wyoming. Their findings suggested that utilizing a combination of multiple methods is essential for accurately predicting deck conditions [5]. These studies highlight that each NDE technique is efficient for the assessment of certain types of deterioration mechanisms, and that considering more than one NDE technology can be beneficial in obtaining a more comprehensive picture of the condition of the deck [5]–[7]. The main objective of any type of condition assessment, such as nondestructive evaluation (NDE), is to facilitate the modeling of deterioration over time, enabling the prediction of its future progression and estimation of remaining lifespan. Traditionally, this has involved converting NDE findings into a quantitative composite index, which is then used to estimate an appropriate deterioration function through regression. Gucunski and collaborators determined a condition index for the bridge deck by calculating a weighted average of deck areas with varying NDE measurements. They utilized this index to fit a sigmoid function for predicting future condition [8]. Kim and colleagues expanded on this research by computing condition indices for subsections of the deck, leading to the creation of spatial condition index maps [9]. Subsequently, the researchers utilized the parameters derived from the fitted sigmoid function to predict condition index maps for different time points. Rashidi et al. introduced a deterioration metric derived from the square root of the Shannon Jensen distance of the probability distribution of NDE data. They applied this metric to develop deterioration models that elucidate the time-dependent behavior of a bridge deck [2]. Babanejad et al. introduced a multi-step framework for estimating the service life expectancy of bridges, which includes incorporating observations from nondestructive evaluation (NDE) [10]. Alsharqawi et al. introduced a method for assessing bridge deck conditions utilizing the Quality Function Deployment (QFD) theory. This approach enables the integration of nondestructive evaluation (NDE) measurements with visual inspection findings by computing a weighted average of severities associated with various defect modes [11]. Ghodoosi et al. integrated Ground Penetrating Radar (GPR) data into a reliability-based model for bridge deck deterioration. This integration involved incorporating defect maps generated from GPR into the calculation of structural reliability factors [12].

### *1.2. Research Gaps: Detailed condition forecasting through spatial-temporal- learning*

While advancements in NDE technologies have improved data collection, further investment in NDE data analytics is crucial to incorporate state-of-the-art artificial intelligence (AI) frameworks [13], [14]. Integrating advanced AI frameworks into NDE analytics shows promise in significantly enhancing data interpretation, prediction, and overall usefulness [15]. The conventional approach in NDE condition assessment often relies on simple mathematical models, like sigmoid functions, to represent infrastructure element conditions over time. These models are typically used to generate a global condition index reflecting the overall state of a component, such as a bridge deck. However, this approach has limitations, especially in capturing the nuanced progression of localized deterioration phenomena like delamination, corrosion, or patchwork in concrete structures. These limitations arise from the global nature of condition indices, which may not capture the dynamics of deterioration processes adequately.

### *1.3. The Need for Advanced Data Analytics*

The integration of cutting-edge AI and machine learning technologies, including Physics-Informed Neural Networks (PINNs) and other spatial-temporal models, into NDE data analytics presents a promising solution to these limitations. Such frameworks have the potential to revolutionize the field by enabling more granular, accurate, and predictive analyses of infrastructure conditions. By leveraging spatial-temporal learning, it becomes possible to not only identify and characterize local deterioration phenomena with high precision but also to forecast their progression over time with a level of detail previously unattainable through traditional models [16] .

The identified research gap underscores the necessity for a paradigm shift in how NDE data is analyzed and interpreted. By embracing the advancements in AI and machine learning, particularly through the application of spatial-temporal learning frameworks, the field of NDE can significantly advance its capabilities in detailed condition forecasting. This shift promises not only to enhance the precision and reliability of infrastructure assessments but also to contribute to the development of more sustainable, resilient, and safe built environments.

### *1.4. Artificial Intelligence for Condition Assessment of Infrastructure*

Recent progress in artificial intelligence, particularly in machine (deep) learning, has facilitated significant advancements in big data processing and the creation of intelligent, autonomous, and interconnected systems [17]. In the field of civil engineering, the sheer volume and intricacy of data related to the built environment suggest an untapped potential for integrating these technologies. In this context, automated image-based approaches for condition assessment, particularly those utilizing convolutional neural networks (CNN), have recently garnered considerable attention in the infrastructure research community [18]–[20].

### *1.5. Spatiotemporal Convolutional Recurrent Neural Networks*

While convolutional neural networks (CNNs) are adept at extracting spatial patterns from image data, Long Short-Term Memory (LSTM) neural networks have been developed to capture temporal features from sequential and time-series data. An LSTM comprises units, each containing a memory cell to retain historical information and gates that regulate information flow. These units are connected successively to form a directed graph suitable for processing temporal sequences. LSTMs have proven highly effective for sequence modeling and time series forecasting, with inputs consisting of sequences of one-dimensional vectors focusing on temporal structures in the data. However, for sequences of image data, combining CNNs and LSTMs enables the incorporation of both the spatial features of the 2D (e.g., image) data and the temporal information of the sequence [21], [22]. Convolutional LSTMs (ConvLSTM) represent an alternative approach to the task of modeling and forecasting image sequences [23]. Convolutional LSTMs (ConvLSTM) integrate the convolution operation into each LSTM unit, enabling them to capture spatial structures within images. ConvLSTM has been applied to forecast short-term precipitation based on sequences of radar echo maps [23]. In this research, the input and output consisted of sequences of B-scan Ground Penetrating Radar (GPR) data. The network was trained to predict future maps based on patterns in previous ones. Zhang et al. employed ConvLSTMs to forecast tumor growth using a dataset of computerized tomography (CT) images from 33 patients with pancreatic tumors [24]. There was a series of three CT scans per patient, with an average time interval of less than two years between consecutive scans. Other past applications of ConvLSTMs include forecasting vibration measurements properties [25] , urban land use maps [26], and traffic flow [27].

## 2. Physics Informed Convolutional Neural Networks

The obstacle in relying solely on model-driven approaches, such as those previously discussed, lies in their substantial data requirements for training, which presents a significant hurdle due to the difficulty in conducting numerous tests to amass this data. While some studies have turned to simulation software to create the needed datasets, the resulting simulated data often lacks realism and fails to accurately mirror actual data. Furthermore, these model-driven models struggle with generalization to data they have not previously encountered. Without any intrinsic comprehension of the data's underlying physical principles, there is no assurance that the models will produce relevant predictions, as their training is purely data dependent.

Within this context of diverse and complementary NDE techniques, Physics Informed Neural Networks (PINNs) represent a novel computational framework that integrates physical laws into the learning process of deep neural networks, offering significant potential for enhancing the predictive accuracy and interpretability of GPR data analysis. This integration promises to revolutionize the way NDE data, particularly from GPR, is interpreted and used for the condition assessment and life expectancy estimation of concrete structures, by providing a more nuanced understanding of subsurface conditions and deterioration mechanisms.

PINNs are built on a foundation that merges the principles of physics with the capabilities of deep learning. By incorporating governing physical laws (e.g., Maxwell's equations for electromagnetic wave propagation) directly into the neural network's architecture, PINNs ensure that predictions are not only data-driven but also constrained by established physical theories [16], [28]. This dual reliance on empirical data and theoretical knowledge allows for the development of models that are both accurate and interpretable, addressing a common critique of traditional deep learning models as "black boxes."

### *2.1. Proposed Research*

In this study, Physics-Informed Neural Networks are combined with data-driven (deep learning) models to capitalize on the physical characteristics inherent in the data, which are implicitly contained within it. The integration of Physics-Informed Neural Networks into the analysis of NDE data, particularly GPR, represents a significant leap forward in the field of structural health monitoring and assessment. By ensuring that predictions are grounded in both data and physical laws, PINNs offer a path towards more accurate, reliable, and interpretable analysis of concrete structures. As this technology matures, it is expected to play an increasingly central role in the condition assessment and life expectancy estimation of infrastructure, ultimately contributing to safer and more sustainable built environments.

The suggested research concept utilizes existing NDE (Non-Destructive Evaluation) data, incorporating NDE surveys gathered via the BEAST program, to assess the viability of the proposed approach. The processed BEAST data that can be found on the https://infobridge.fhwa.dot.gov/ website. The idea involves employing time-series NDE maps in conjunction with ConvLSTM and attention layers, taking advantage of the spatial and temporal coherence in NDE time series to predict future conditions of NDE maps. The approach treats GPR (Ground Penetrating Radar) raw data maps as a problem of forecasting spatial-temporal sequences.

This work incorporates information from a temporal sequence of raw images in addition to the spatial consistency information. It should be noted that while the feasibility and superiority of creating forecasting models based on recurrent neural networks will be studied and demonstrated in this proposed research, such models will continue to improve as more data will be available through NDE data collection programs.

*2.2. Physics Informed Neural Network – Objective Function details.*

Ground-penetrating radar (GPR) utilizes electromagnetic waves to image the subsurface of the ground. The fundamental electromagnetic wave propagation is governed by the wave equation, which can be described as [29], [30]:

$$\nabla^2 E - \mu\varepsilon \frac{\partial^2 E}{\partial t^2} - \mu\sigma \frac{\partial E}{\partial t} = 0 \tag{1}$$

where $\mu$ is the permeability of the medium and $\varepsilon$ is the permittivity of the medium. This equation considers both the propagation and attenuation of the wave due to the conductive medium, and thus is more realistic for GPR applications which often involve penetration through various materials with different conductivities and permittivities [29]. In the context of GPR, solving this equation with appropriate boundary conditions can simulate how radar waves propagate through and are reflected by different subsurface structures. To integrate a Physics-Informed Neural Network (PINN) into the Ground Penetrating Radar Technique simulation, one need to embed the knowledge of the physical equations governing the wave propagation into your neural network structure.

Let us define $f(t, x)$ to be given by

$$f \coloneqq \nabla^2 E - \mu\varepsilon \frac{\partial^2 E}{\partial t^2} - \mu\sigma \frac{\partial E}{\partial t} \tag{2}$$

and proceed by approximating $E(t, x)$ by a deep neural network. To this end, $E(t, x)$ can be simply defined as

```
def E(t , x ) :
    u = deep_net ( t f . layers ( [ t , x ] , 1 ) , weights , biases )
    return u
```

Correspondingly the physics-informed deep networks $f(t, x)$ takes the form [31], [32]:

```
def f ( t , x ) :
    E = E( t , x )
    E t = t f . gradients (E , t ) [ 0 ]
   E tt = t f . gradients (Et , t ) [ 0 ]
   E x = t f . gradients (E , x ) [ 0 ]
```

```
E xx = t f . gradients ( E x , x ) [ 0 ]
f = Exx- μεEtt- μσEt
return f
```

Every parameter in Equation 1 must either be predefined or designated as a trainable variable that is adjusted and optimized according to the training data. Given that the structure is assumed to be dry concrete, the value of $\sigma$ is set to zero. Consequently, Equation 1 can be reformulated as follows:

$$\nabla^2 E - \mu\varepsilon \frac{\partial^2 E}{\partial t^2} = 0 \quad (3)$$

To determine the dielectric parameter referenced in Equation 2, two main steps are followed: initially, the value of the dielectric parameter is estimated based on existing experimental assumptions and the nature of the structure in question. Subsequently, this parameter is designated as a trainable variable, allowing it to be refined during the training process based on the specific data being used. This approach enhances the model's accuracy by enabling the selection and adjustment of material properties to better align with real-world data and conditions.

The shared parameters between the deep networks $E(t,x)$ and $f(t,x)$ can be learned by minimizing the mean squared error loss:

$$MSE = MSE_E + MSE_f \quad (4)$$

where

$$MSE_u = \frac{1}{N_E} \sum_{i=1}^{N_E} |E(t_E^i, x_E^i) - E^i|^2 \quad (5)$$

and

$$MSE_f = \frac{1}{N_f} \sum_{i=1}^{N_f} |f(t_f^i, x_f^i)|^2 \quad (6)$$

Here, $\{t_E^i, x_E^i, E^i\}_{i=1}^{N_E}$ denote the training data on $E(t,x)$ and $\{t_f^i, x_f^i\}_{i=1}^{N_f}$ specify the data for $f(t,x)$. $MSE_E$ corresponds to the error function of energy based on deep networks and $MSE_f$ enforces the structure imposed by equation 1 at a finite set of colocation points. To determine the percentage impact of both $MSE_E$ and $MSE_f$, Equation 4 is expanded to include a weight parameter known as $\alpha$.

$$MSE = \alpha MSE_E + (1-\alpha) MSE_f \quad (7)$$

The parameter α is chosen as a trainable variable, which is optimized throughout the training process to enhance the model's performance based on specific criteria.

In our work, we replace $E(t,x)$ by a deep network $E(t,x;W,b)$ and obtain a physics-informed neural network $f(t,x;W,b)$ by automatic differentiation. Consequently, the resulting pair $E(t,x;W,b)\ and\ f(t,x;W,b)$ must satisfy the fundamental electromagnetic wave propagation regardless of the choice of the weights $W$ and bias $b$ parameters. During training, given a data-set $t_i, x_i, E_i$, and $t_j, x_j, f_j$ we then try to find the correct parameters $W^*$ and $b^*$ such that we get as good a fit as possible to both the observed data and the differential equation residual.

In all benchmarks considered in this work, the total number of training data $N_E$ is relatively small (a few hundred to a few thousand points) and we chose to optimize all loss functions using ADAM optimization algorithm. Despite the fact that there is no theoretical guarantee that this procedure converges to a global minimum, our empirical evidence indicates that, if the given partial differential equation is well-posed and its solution is unique, our method is capable of achieving good prediction accuracy given a sufficiently expressive neural network architecture and a sufficient number of collocation points $N_f$. The training concept of the physics-informed neural network is depicted in Figure 2.

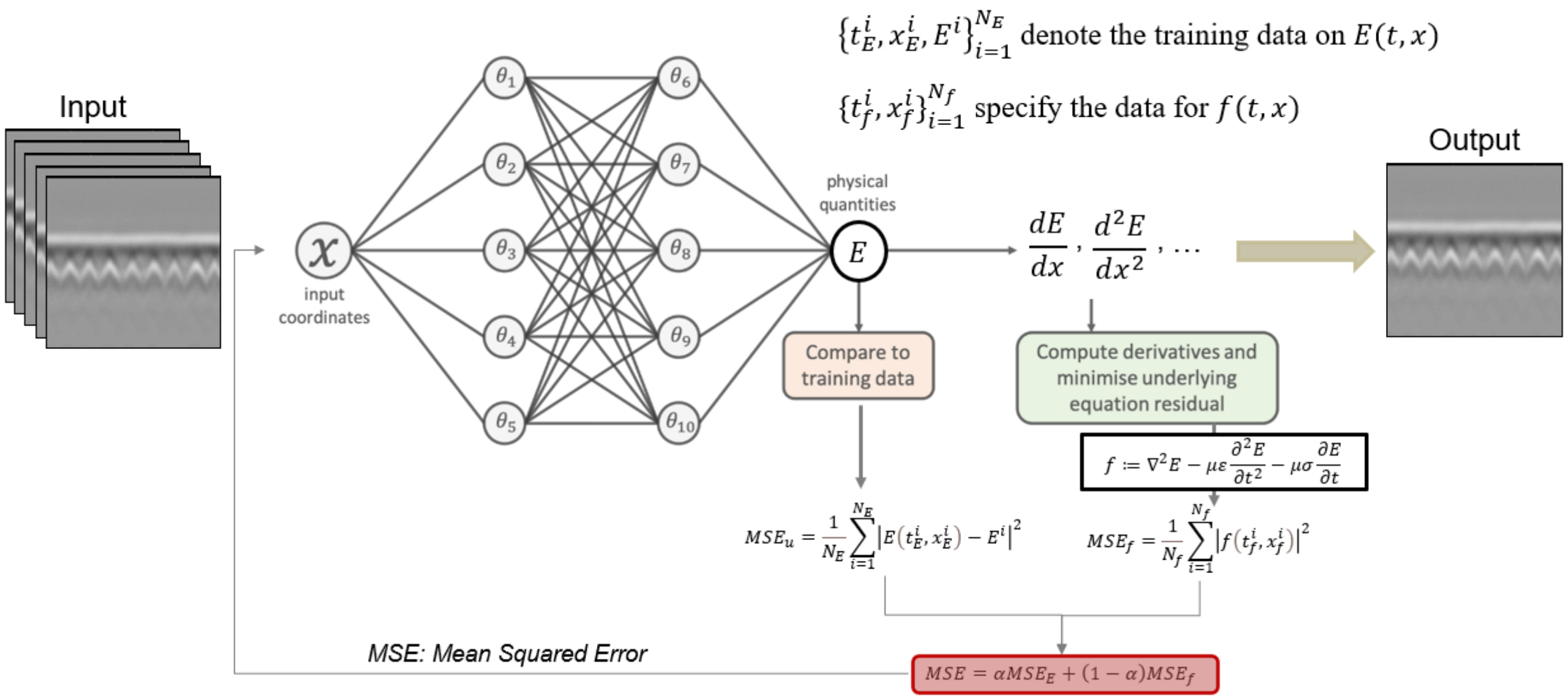

Figure 2. Concept training of the physics informed neural network.

## 3. Deep Network Model Architecture

The proposed deep learning architecture consists of three parts: CNN as detailed, a spatial feature channel attention (SFCA) mechanism, ConvLSTM, followed by temporal feature frame attention (TFFA) modules which are described in the following sections. The attention mechanism calculates the channel attention and temporal attention weights with self-adaption and adjusts the visual and temporal features response to extract the most relevant and salient visual and temporal features (Figure 3).

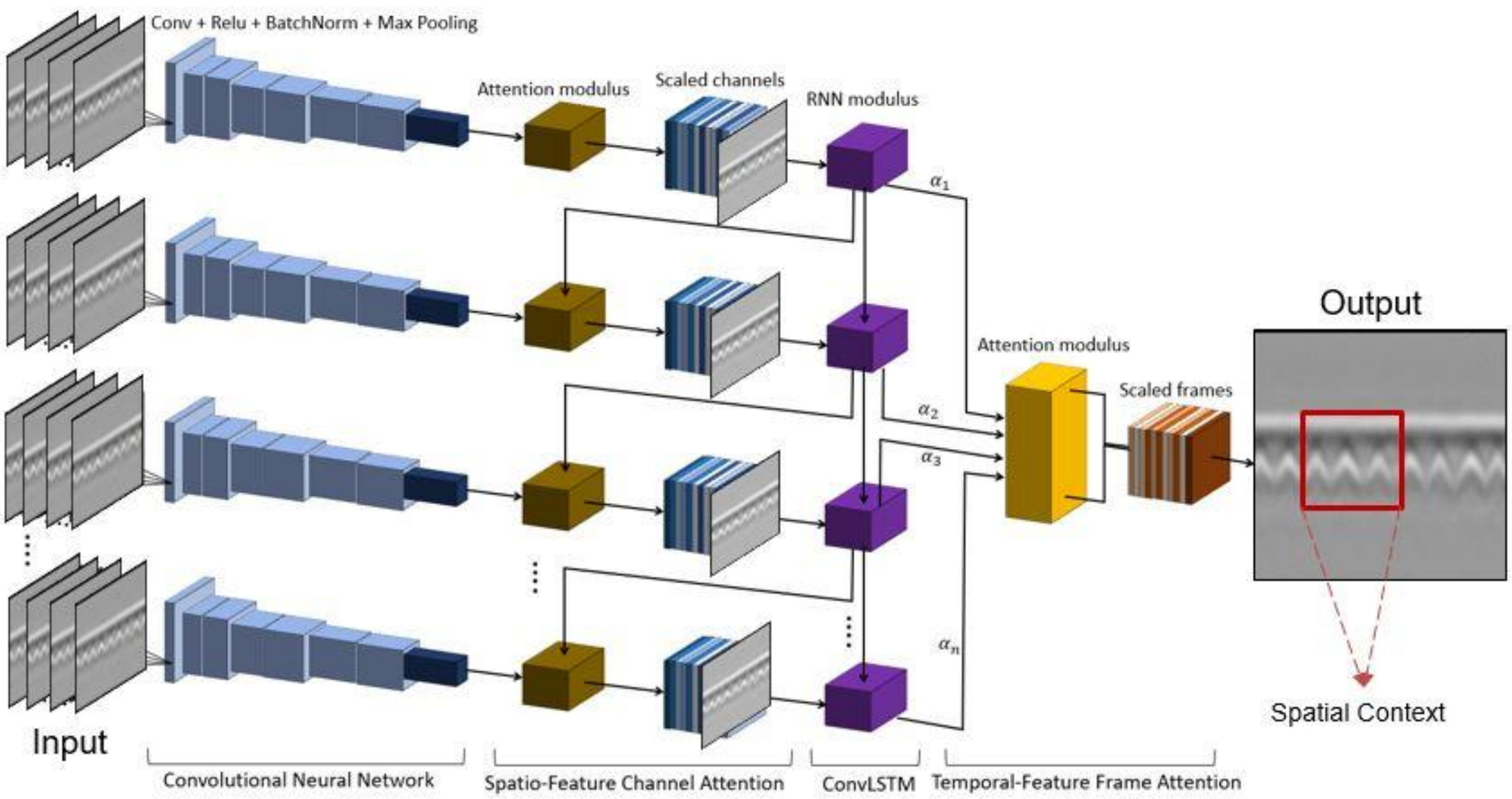

Figure 3. A detailed flow of attention blocks in the DLVM. Brief overview of the DLVM architecture showing spatial feature channel attention (SFCA), ConvLSTM and temporal feature frame attention (TFFA).

### *3.1. CNN Architecture in PINN*

The components of the proposed CNN include one input layer, two inception blocks, two convolution layers, one max-pooling layer, an average-pooling layer, and one regression layer as the output of the network. The proposed CNN used in this study to recognize spatial features is illustrated in Figure 4. Sensitivity analysis may be applied to the number of the convolution layers as well as the values for *f, d*, and *s* to find the optimal values; this is initially conducted in this work, and the optimized CNN is utilized in the model. The parameter *f* for each layer is the number of output feature maps in the previous layer (the input layer passes on only one feature map). The two-convolution layers are followed by a max-pooling layer, which down-samples the feature maps along the spatial dimension. The parameters of the pooling layers have the dimension of the block ($m \times m \times m$) and the spatial stride (*s*). According to the common practice among the research community, a value of 2 is adopted in this work for both *m* and *s*.

To provide more detailed information and extract patterns with different sizes, filters (convolutions) with varying sizes are used at the same convolutional layer; importantly, a 1 × 1 convolutional layer, the so-called bottleneck layer, is employed to decrease both the computational complexity and the number of parameters. To be more precise, the 1 × 1 convolutional layer is adopted just before a larger kernel convolutional filter (e.g., 3 × 3 and 5 × 5 convolutional layers) to decrease the number of parameters to be determined at each level (i.e., the pooling feature process). In addition, the 1 × 1 convolutional layer makes the network deeper and adds more nonlinearity by using *ReLU* after each layer. In this network, the fully-connected layers are replaced with an average pooling layer, which significantly decreases the number of parameters since the fully-connected layers include a large number of parameters. Thus, as sketched in Figure 4, this network can learn deeper representations of features with fewer parameters relative to AlexNet or ResNet while it is much faster than VGG.

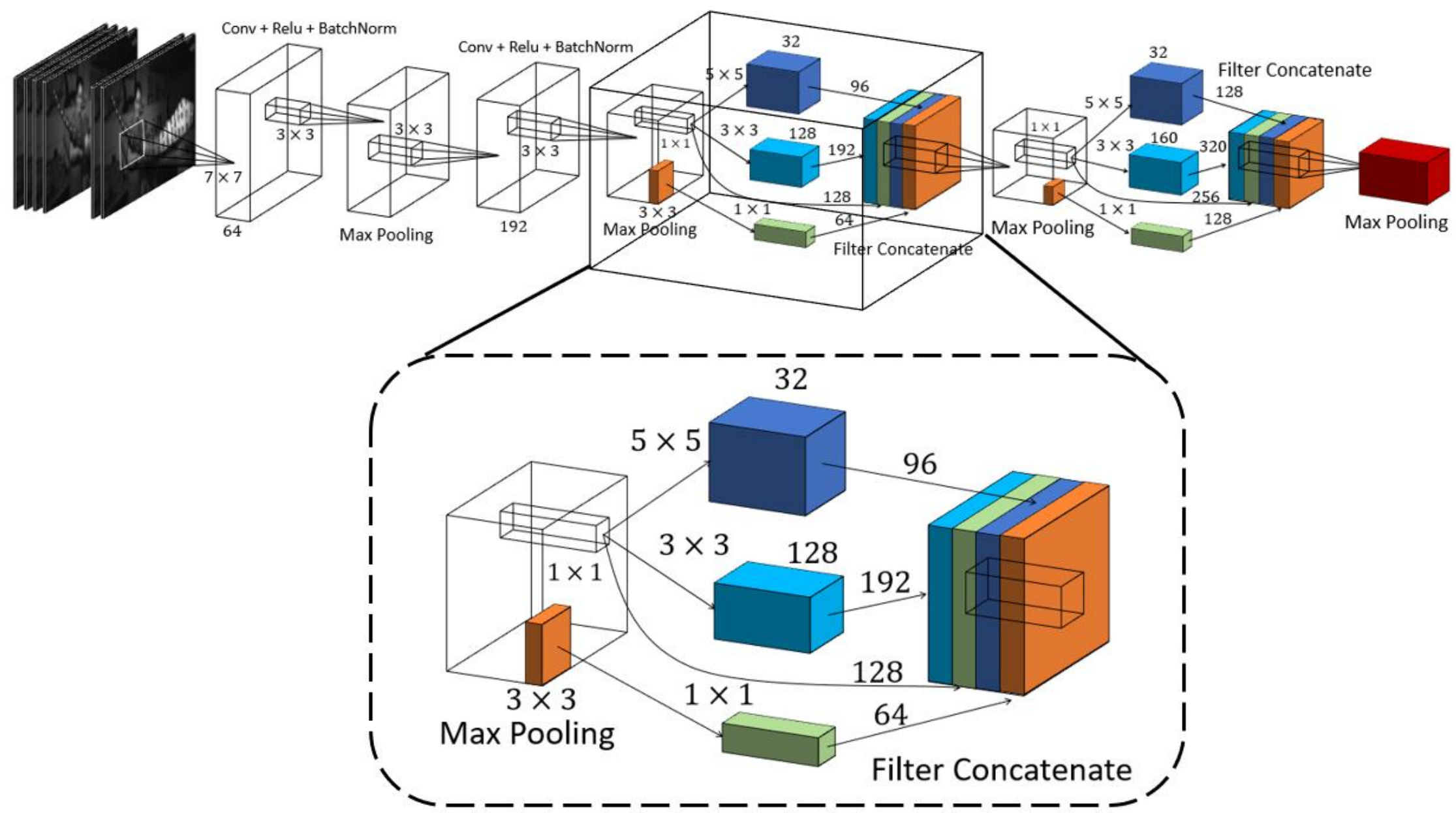

Figure 4. A comprehensive detail of the CNN module in the proposed PINN

*3.2. ConvLSTM Module in PINN*

To better explore the correlation among frames, this paper considers the prediction as a sequential task. To thoroughly explore the correlation of temporal variables in the process of sequential prediction, the suggested model detects the connections between the existing frames one at a time. Recurrent neural networks (RNN), which were created expressly for processing sequence data samples, not only propagate each layer to the next but also output a hidden state for the current layer when processing the subsequent sample. Long short-term memory (LSTM), an evolutionary variant of RNN that addresses the issue of gradient expansion and gradient disappearance in RNNs, is good at mining the correlation between sequential samples. LSTMs can make use of the hidden state containing historical data for forecasting future qualities. LSTMs excel at sequence modeling tasks, however, when processing images, the general LSTMs disregard spatial information. This is because general LSTMs flatten the input image into a one-dimensional vector and model the sequence information through the whole connection layer, which results in the loss of image spatial information and is not advantageous for the advancement of GPR prediction. For GPR data recognition, different image areas have different levels of importance for different feature predictions. Therefore, retaining relevant spatial information will be very helpful for improving performance. To maintain the spatial structure of the images, convolutional LSTM (ConvLSTM) [33]–[35] is utilized in this work instead of a standard LSTM. It can capture the spatial information of images better than standard LSTM, and could, therefore, better mine the correlation between the images. The equations for ConvLSTM is established as follows:

$$i_t = sigmoid(W_{ix} * x_t + W_{ih} * h_{t-1} + W_{ic} \odot c_{t-1} + b_i) \quad (8)$$

$$f_t = sigmoid(W_{fx} * x_t + W_{fh} * h_{t-1} + W_{fc} \odot c_{t-1} + b_f) \quad (9)$$

$$o_t = sigmoid(W_{ox} * x_t + W_{oh} * h_{t-1} + W_{oc} \odot c_t + b_o) \quad (10)$$

$$g_t = tanh(W_{gx} * x_t + W_{gh} * h_{t-1} + b_g) \quad (11)$$

$$c_t = f_t \odot c_{t-1} + i_t \odot g_t \quad (12)$$

$$h_t = o_t \odot \tanh(c_t) \quad (13)$$

where * contributes to the convolution operation and $\odot$ accounted for element multiplication, *sigmoid*(.) *is* regarded as the logistic sigmoid function and *tanh*(.) represents the hyperbolic tangent function, the script *t* represents the *t-th* step of ConvLSTM, $i_t$ is input gate, $f_t$ is forget gate, $o_t$ is output gate, $g_t$ is input modulation gate, $x_t$ is input data, $c_t$ is the cell state, $h_t$ is the hidden state. Here, $x_t, c_t, h_t, i_t, f_t, o_t$ are three-dimensional tensors; the first dimension expresses temporal information, and the second and third dimension expresses the rows and columns of spatial information. The convolution operation is utilized here to keep the spatial information of the images. In experiments conducted in this work, ConvLSTM can achieve better performance compared to standard LSTM due to its focus on the key areas with more feature significance. The internal structure of ConvLSTM applied in the *PINN* model is shown in Figure *5*. As it can be noted, Figure *5*a shows the details of one ConvLSTM block, and Figure *5*b shows how those blocks are connected to exchange the generated information and passing previous information to the next block.

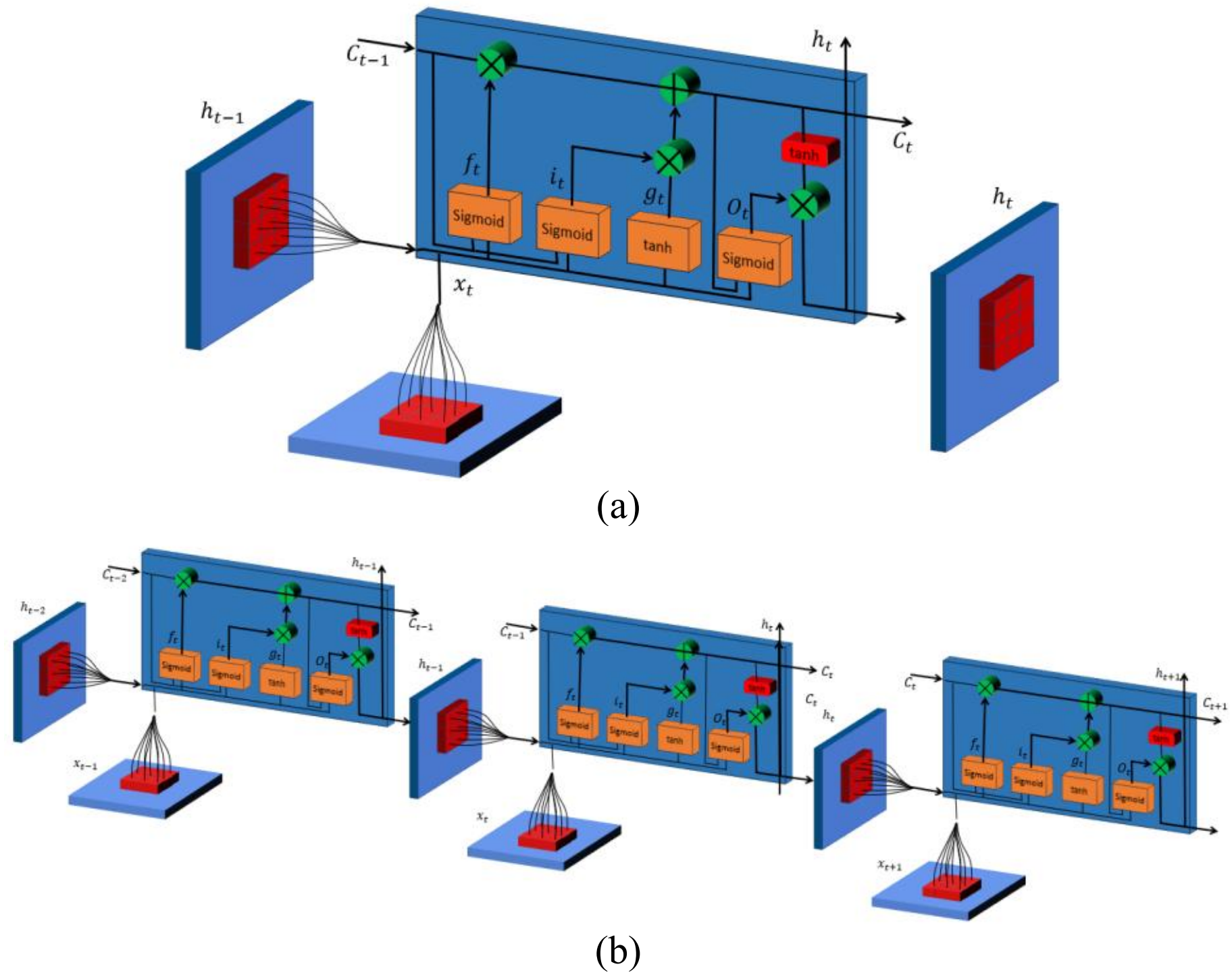

(a)

(b)

Figure 5. The overview of the ConvLSTM, (a) The details of the ConvLSTM with its components, (b) The connected ConvLSTM to exchange information block by block.

*3.3. Spatial-Feature Channel Attention (SFCA)*

Implicit spatial attention is a result of the ConvLSTM mentioned in the preceding section. Additionally, it can concentrate on crucial feature-related areas and significantly enhance GPR data identification performance. The role of Spatial-Feature-Channel Attention (SFCA) is to increase the feature representation capability of the networks by modeling the dependence of each channel in the feature map [33]–[35]. Through global pooling in the channel dimension, SFCA first gains the global information of each channel. To get the aim of feature response and recalibration, it then adaptively models the correlation between channels and weights each channel according to the correlation. This allows the network to selectively improve features that convey relevant information while suppressing aspects that are pointless or ineffectual. In this study, a channel-based attention mechanism is developed to carry out the task of GPR data identification using more recognizable and pertinent information. The spatial attention of ConvLSTM is merged with the SFCA to additionally boost recognition performance. The details of the SFCA mechanism are illustrated in Figure *6*. In this work, SFCA is integrated into the ConvLSTM sequential prediction process, and the input image features and ConvLSTM hidden states are fused for SFCA weight computation. When predicting various features, ConvLSTM will incrementally forecast the important dependency of temporal features and adaptively change the weight of various feature responses.

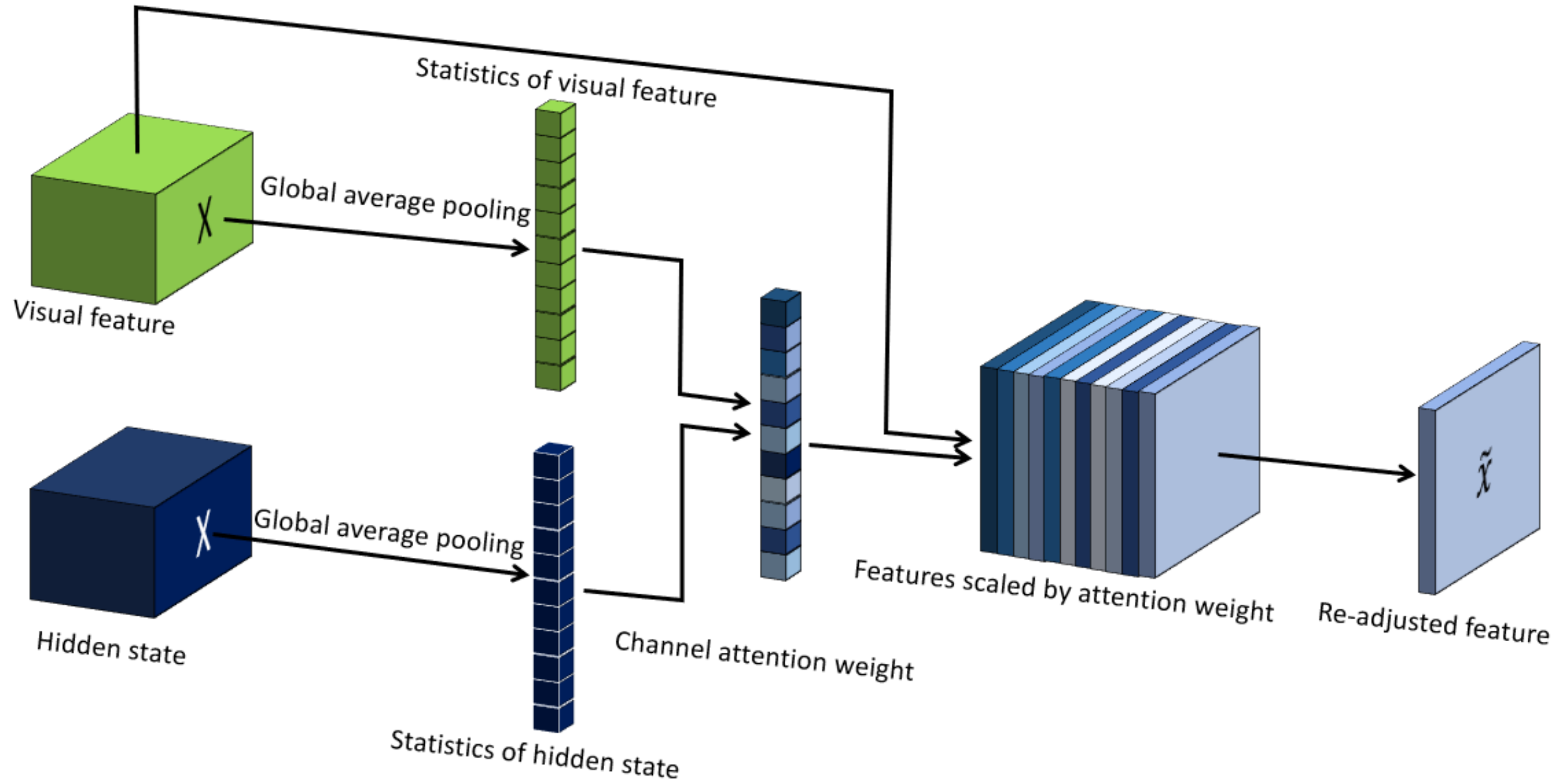

Figure 6. Spatial-feature channel attention blocks.

The objective of this work was to determine the attention weight of each feature response channel by using global average pooling (GAP) to obtain the matching statistical data of the visual features of each channel as the description of global spatial information based on the channel. Additionally, the hidden state of ConvLSTM is considered in the channel-based statistics to adaptively gain the SFCA weight based on the previously predicted attribute. The equations of these two categories of statistical information are as follows:

$$m_c = g(x_c) = \frac{1}{W \times H}\sum_{i=1}^{W}\sum_{j=1}^{H} x_c(i,j) \tag{14}$$

$$n_c = g(h_{t-1,c}) = \frac{1}{W \times H}\sum_{i=1}^{W}\sum_{j=1}^{H} h_{t-1,c}(i,j) \tag{15}$$

where $g(\cdot)$ accounts for the GAP function, $c$ indicates the *c-th* channel, $x_c$ and $m_c$ represent the visual feature and its statistics, $h_{t-1,c}$ and $n_c$ explains the previous hidden state and its statistics, and $W$ and $H$ represent the width and height of the visual feature. Then, the calculation formula of the channel attention weights is established as follows:

$$z_c = sigmoid(w_2 ReLU(w_1[m_c, n_c] + b_1) + b_2) \tag{16}$$

Ultimately, the input feature is multiplied by the attention weight of each channel to get the re-adjusted feature:

$$\tilde{x} = \sum_{c=1}^{C} z_c x_c \tag{17}$$

*3.4. Temporal-Feature Frame Attention (TFA)*

The previous section uses SFCA to optimize filters corresponding to spatial channels, paying attention to essential features for future of GPR data extraction. This section introduces Temporal-Feature Frame Attention (TFFA) to focus on critical frames in the sequences of images [36]. Since our architecture's input is sequences of images, the variations between the frames encode additional helpful information to differentiate GPR data features. To capture the temporal dynamic information across frames, the ConvLSTM architecture is employed to consider sequences of CNN activations explicitly. The primary assumption in sequence modeling networks such as RNNs, LSTMs, and GRUs is that the current state holds information for the whole input observed so far. Hence, after reading the whole input sequence, the final state of an RNN should contain complete information about that sequence, which is often a too strong condition. TFFA mechanism relaxes this assumption and proposes looking at the hidden states corresponding to the whole input sequence to make any prediction. TFFA mechanism concept is demonstrated in Figure *7* conceptually.

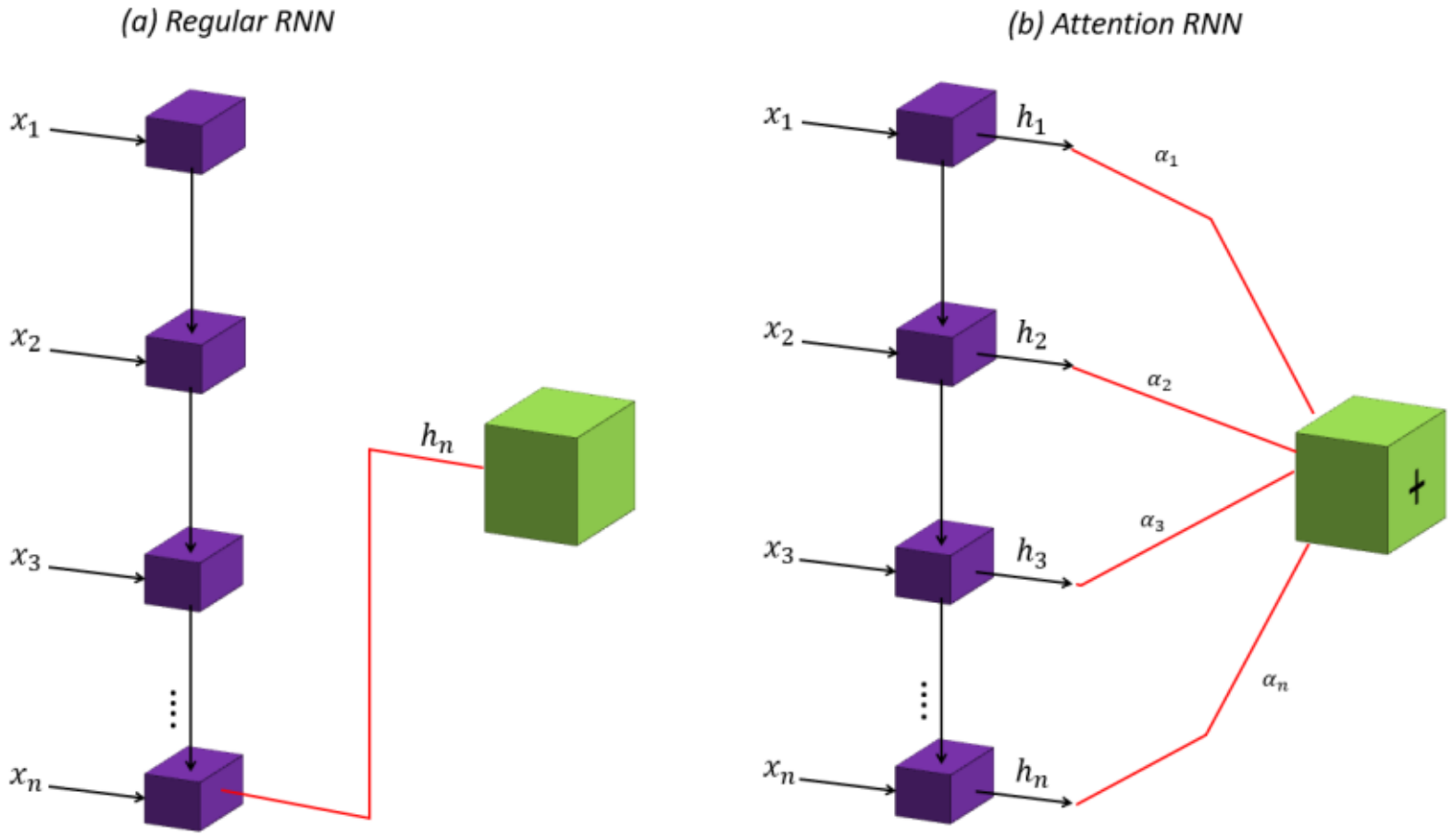

Figure 7. (a) Regular RNN, (b) Temporal Attention based RNN.

The TFFA mechanism is introduced to the ConvLSTM network to weight keyframes in the corresponding videos, and along with the training of ConvLSTM, the hidden sequences are encoded into a fixed-length vector *c* as the following equation:

$$c = \sum_{i=1}^{N} \alpha_i h_i \qquad i \in 1, \dots, N \tag{18}$$

where $h_i$ is the hidden state at a time $t_i$, and the weight $\alpha_i$ is the corresponding weight mapping $h_i$ to vector *c*. At each time step, $\alpha_i$ is computed by

$$\alpha_i = \frac{\exp(W_i^T h_{i-1})}{\sum_{j=1}^{N} \exp(W_j^T h_{i-1})} \tag{19}$$

where $\alpha_i$ can be regarded as the probability dictating the importance of the hidden state $h_i$ . This attention mechanism decides which frames from the input sequences must be received attention. Following computing these probabilities, the model would concentrate on the samples with dynamic changes rather than expecting all the input sequences. The fixed-length vector $c$ is eventually attached to the regression layer to calculate the particular probabilities (Figure *8*).

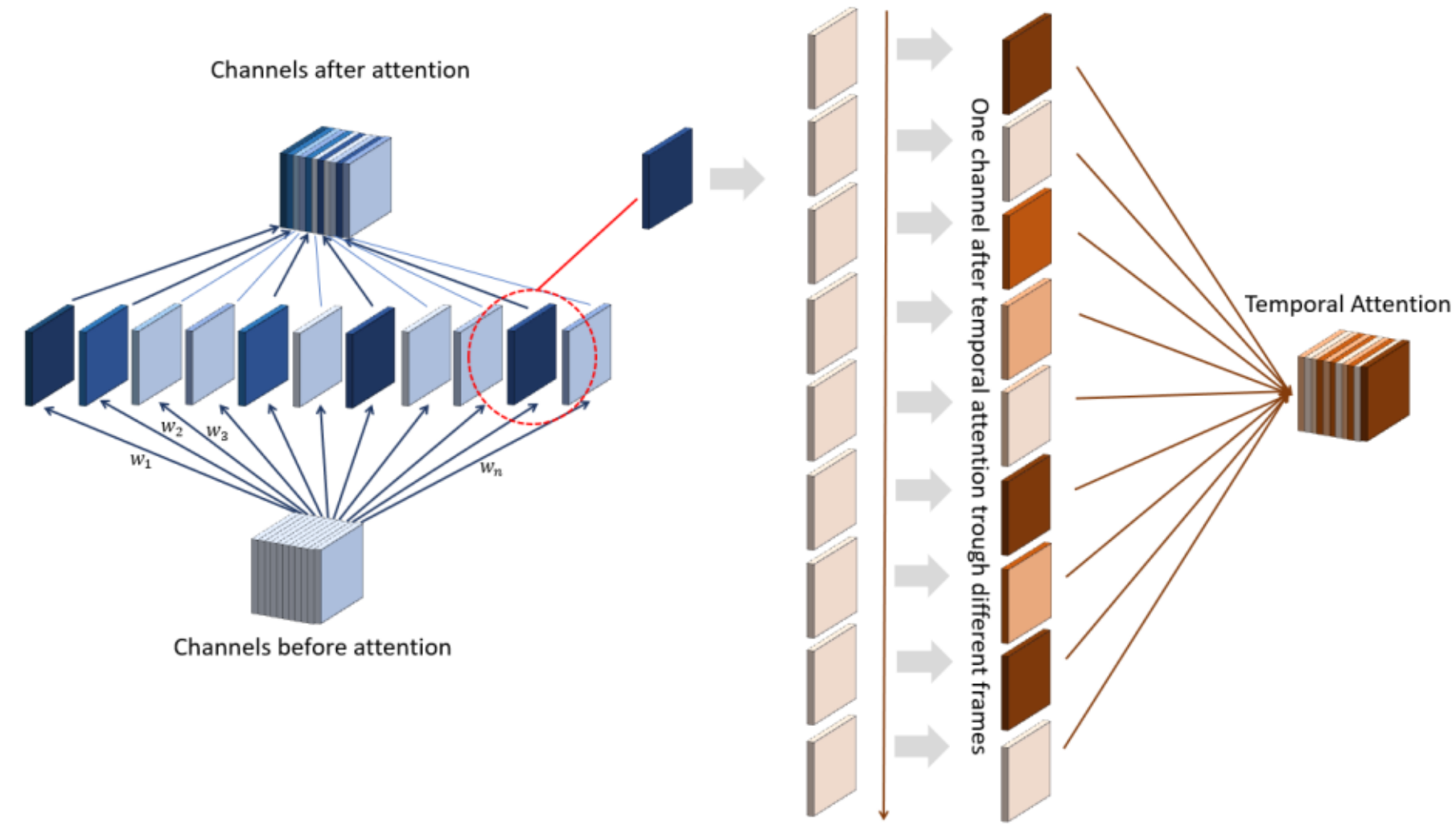

Figure 8. The schematic of spatial channel attention and temporal attention

## 4. Training Details and Network Implementation

The model is trained using Adam, a stochastic optimization algorithm, with $\beta_1 = 0.9$, $\beta_2 = 0.999$, a batch size of 4, and a high weight decay of 0.1 proved to be feasible to transfer model. A linear learning rate warmup and decay are also used. The dataset is divided into three parts: training, validation, and testing. The three categories are 75%, 10%, and 15%, respectively, of the total dataset. In this paper, Mean Squared Error (MSE) is adopted as the evaluation criteria for the proposed deep learning architecture, which is also the loss function for the deep learning architecture during training. MSE is the quantification of the squared error between the prediction and ground truth values and is expressed by Eq. (7). The learned model yields a training and validation MSE of 0.1 and 2.4, respectively, in 400 steps, and its convergence of loss function versus several steps is shown in Figure *9*.

Tensorflow is used to implement the model, and Keras is utilized as the high-level neural network API on top of Tensorflow to speed up the implementation and experimentation. The model is trained from scratch based on *BEAST dataset* shared in https://infobridge.fhwa.dot.gov/ website. Xavier's initialization method is employed in this stage. A previously mentioned Adam optimization approach is used to minimize loss functions where the momentum is set to 0.9. $L2$ regularization is also employed for all weight parameters. The dropout ratio and weight of $L2$ regularization are set to 0.5 and 0.0005 during the entire training process. The learning rate is initialized as 0.0001 and drops by a factor of 10 after the loss is stable. Here the MSE values between predicted and real data for existing data are chosen to evaluate the model performance. The Adam optimizer is employed to train the networks, given its superiority over other stochastic optimization methods. The learning rate is kept at 0.0001 and the total number of training epochs is set to 400. The validation dataset here helps monitor overfitting during training. The network is trained using Keras 2.2.4 (Tensorflow GPU 1.13.1 backend), CUDA 10.0 toolkit, and cuDNN 7.5 support in a commercial laptop, which has 16 GB RAM and a 4GB video RAM GTX 860 M GPU. The details of the deep model architecture are elaborated in Table 1.

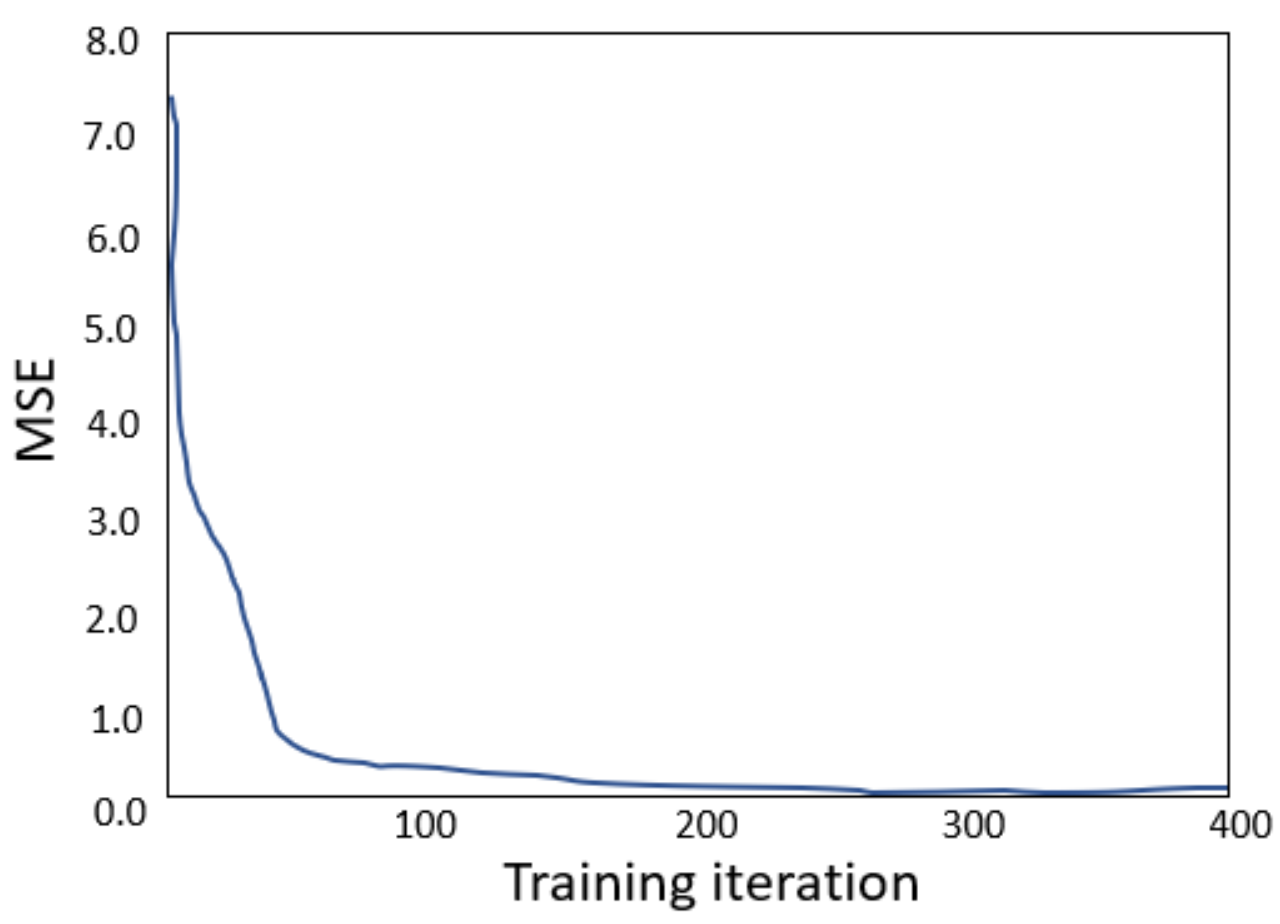

Figure 9. Convergence of loss function with number training steps.

Table 1. The layers configuration of the proposed network

| Layer | Name | Configuration | Modules |
|---|---|---|---|
| 1 | Convolution 1 | 64 kernels, kernel size (7×7) | CNN |
| 2 | Batch Normalization | | |
| 3 | Max Pooling | Pooling size (3×3), Stride (2×2) | |
| 4 | Convolution 2 | 192 kernels, kernel size (3×3) | |
| 5 | Batch Normalization | | |
| 6 | Max Pooling | Pooling size (3×3), Stride (2×2) | |
| 7 | Inception Block 1 | Filters, 128,128,192, 32, 96, 64 | |
| 8 | Max Pooling | Pooling size (3×3), Stride (2×2) | |
| 9 | Inception Block 2 | Filters, 256, 160, 320, 32, 128, 128 | |
| 10 | Max Pooling | Pooling size (3×3), Stride (2×2) | |
| 11 | Attention mechanism | Spatial-Feature Channel Attention | Channel Attention |
| 12 | ConvLSTM | ConvLSTM, 512 hidden unit outputs | RNN |
| 13 | Attention mechanism | Temporal-Feature Frame Attention | Temporal Attention |

*4.1. Discussion on Training Issues and Overfitting*

To avoid overfitting, a less complex CNN model is applied, and the filters are carefully chosen after repeated experimentations with various filter numbers and combinations. Also, adding one fully connected layer and dropouts instead of the proposed one layer gave significant performance improvements with an MSE of 2.3. Increasing the number of ConvLSTM units in the LSTM layer improved the performance a bit, and after that, increasing the number of units does not make a significant difference. Adding a ConvLSTM layer to the network helped improve the network's performance, which enabled the network to learn a more complex time function, and the network improved to an MSE of 0.12.

## 5. Experiment Setup: Bridge Evaluation and Accelerated Structural Testing (BEAST)

BEAST is the first full-scale bridge testing facility located in Piscataway, New Jersey (Figure 10). The BEAST facility is built and commissioned by the Center for Advanced Infrastructure and Transportation at Rutgers, The State University of New Jersey. The BEAST facility is the first facility nationwide capable of applying controlled and accelerated live load, environmental, and maintenance demands on full-scale bridge superstructures. The specimen is a multi-girder steel composite bridge (30 by 50 ft) with an 8-inch bare concrete deck and black rebar reinforcement. It will be subjected to rapid-cycling environmental changes and extreme traffic loading to speed up deterioration, as much as 30 times, in order to simulate 15-20 years of wear-and-tear in just a few months. The deck is supported by four I-beams as the main girders with one fixed and one open joint. At its initial design, the specimen is supposed to be exposed to over 8 million cycles of live loading (60 kips), 400 freeze-thaw and hot-dry cycles, as well as the application of deicing agents (6% brine solution) to simulate common winter maintenance practices. The primary target is to validate performance models by measuring stresses and deterioration caused by live, environmental, and maintenance loading in an extremely compressed time frame. To that end, the BEAST experiment aims to utilize accelerated testing of the full-scale bridge deck and superstructure systems subjected to cyclic moving wheel loads and freeze-thaw environmental conditions. It is primarily envisioned that BEAST experiments will complement both field observations and material-level tests and fill an important gap in our current understanding of bridge performance and deterioration.

The BEAST database encompasses data collected through five Nondestructive Evaluation (NDE) technologies including Impact Echo (IE), Ground Penetrating Radar (GPR), Ultrasonic Surface Waves (USW), Electrical Resistivity (ER), and Half-Cell Potential (HCP). This comprehensive data collection was primarily conducted on the BEAST bridge specimen, focusing on the concrete bridge deck. Data use in this study are the processed BEAST data that can be found on the https://infobridge.fhwa.dot.gov/ website. In this work only GPR data is used to train the model.

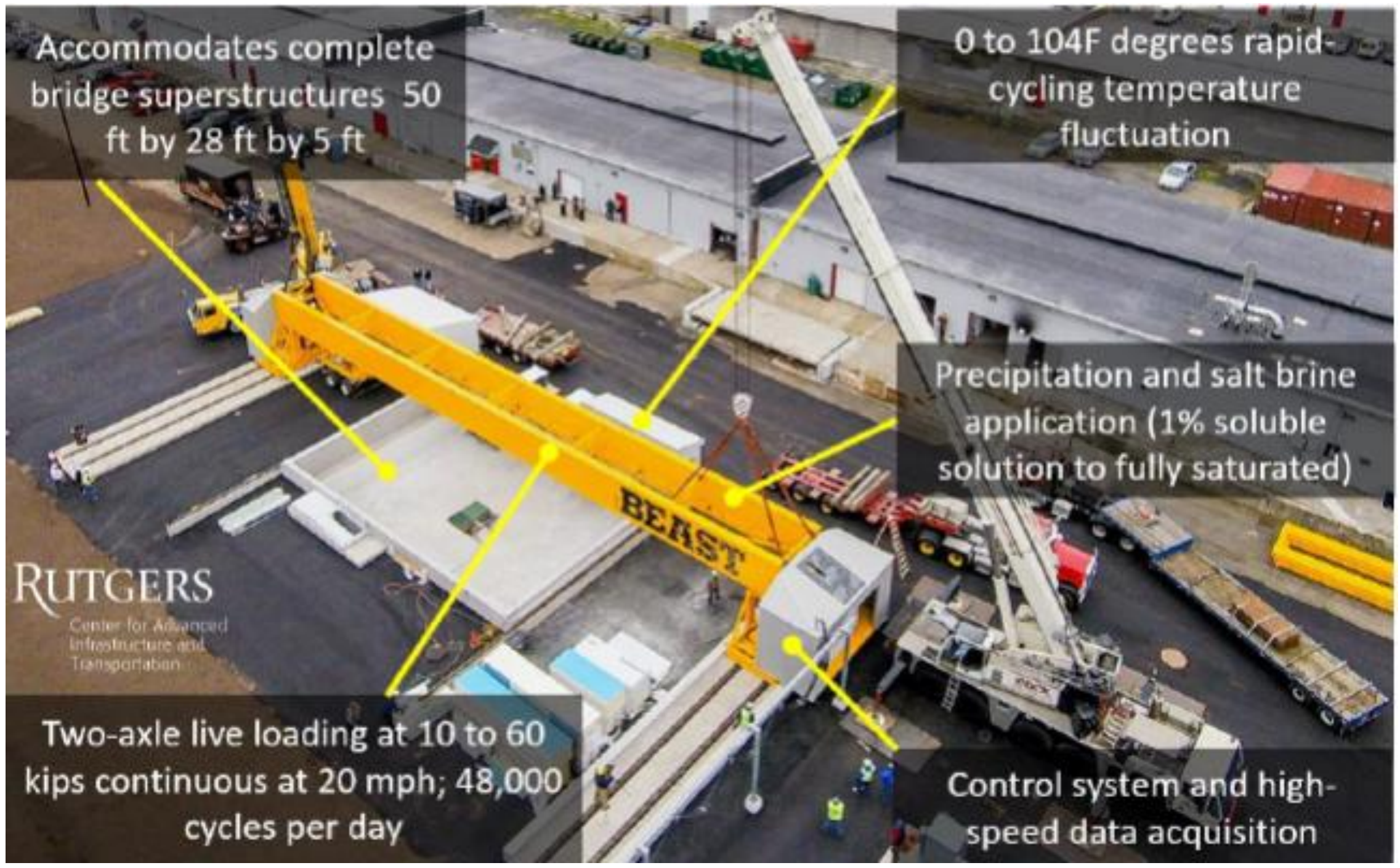

Figure 10.The overview of the BEAST

*5.1. The BEAST data preparation and processing*

Ground Penetrating Radar (GPR) data, sourced from the BEAST initiative, is utilized to train deep learning models known as Physics-Informed Neural Networks (PINNs). This raw data is captured during experimental tests and is stored in the DZT file format. The tests are conducted using a GPR device. To prepare the data for training, it undergoes processing through RADAN software, which is used to refine and format the data suitably. This processing includes cropping and formatting the data into 256 by 256-pixel dimensions. An example of such processed data is displayed in Figure 11, showcasing the specific adjustments made for optimal use in the training phase of the model. This methodological preparation ensures that the data is in the best form to effectively train the neural networks, potentially enhancing the accuracy and efficiency of the model.

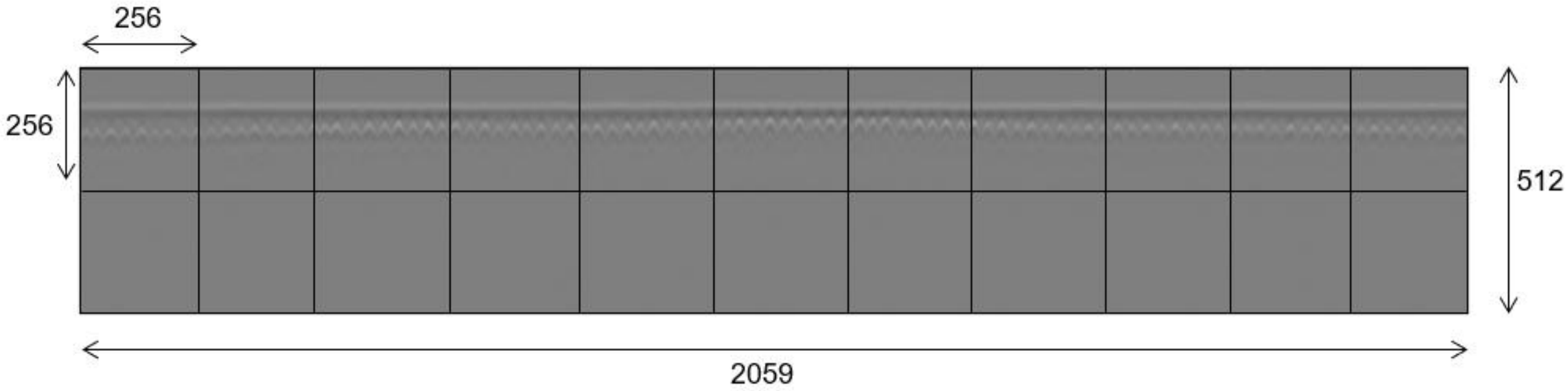

Figure 11. Sample example of BEAST GPR data

As demonstrated in *Figure 11*, the total dimensions of the image are 2051 by 512 data points. Due to the large size of the image, it is sectioned into smaller, cropped images measuring 256 by 256 pixels each for model training purposes. The data is systematically segmented into three distinct sets: "train_set", "test_set", and "validation_set". This division helps in creating a sequence, labeled as *M*, from splitting the original images into *N* smaller images. These segmented images then serve as the sequence of inputs fed into the deep learning engine, aligning with the expected outputs as depicted in Figure *12*a. Also the schematic of training process based on initial data and future dataset is depicted in Figure *12*b. This structured approach not only organizes the data effectively but also optimizes the training process by providing varied data subsets for comprehensive learning and validation of the model.

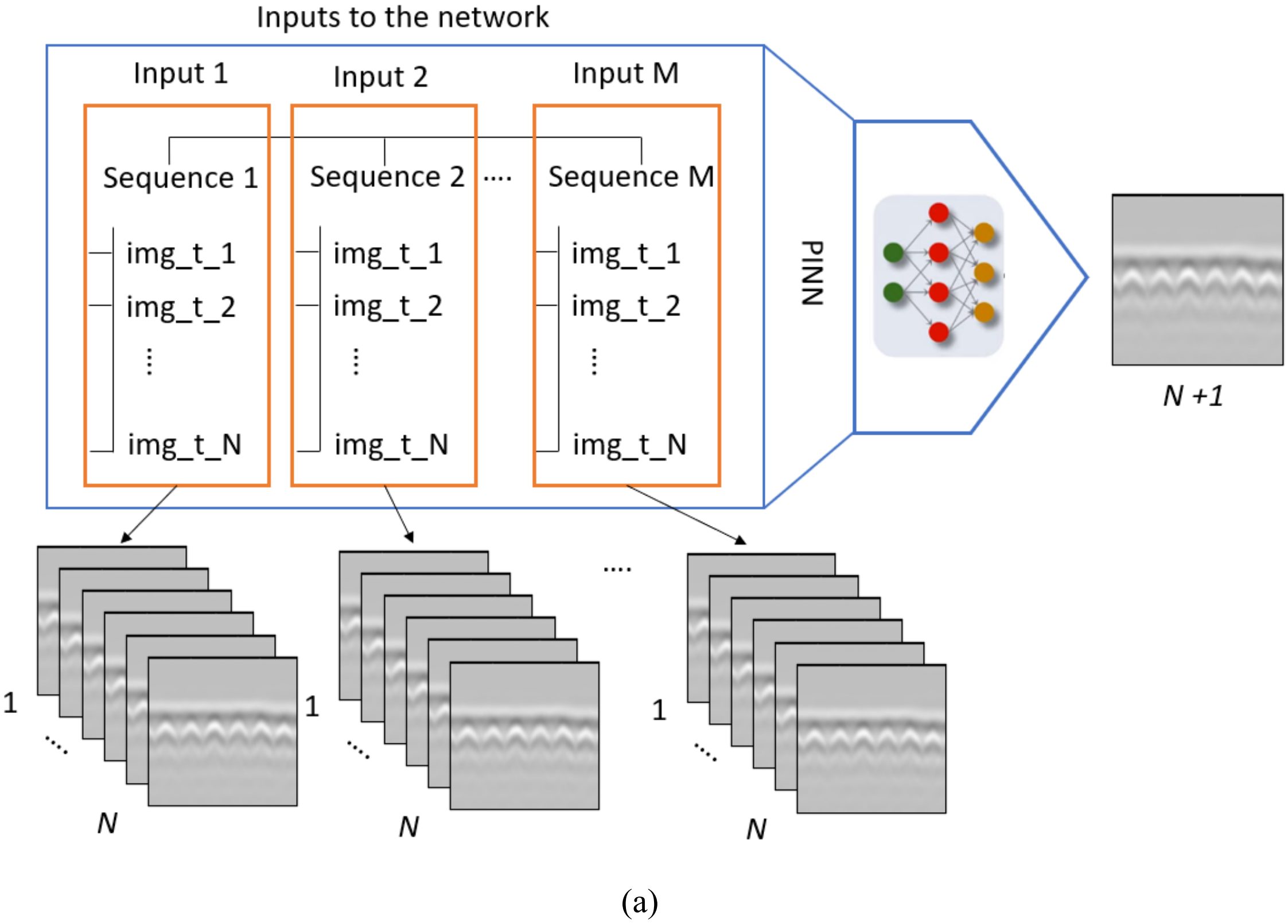

(a)

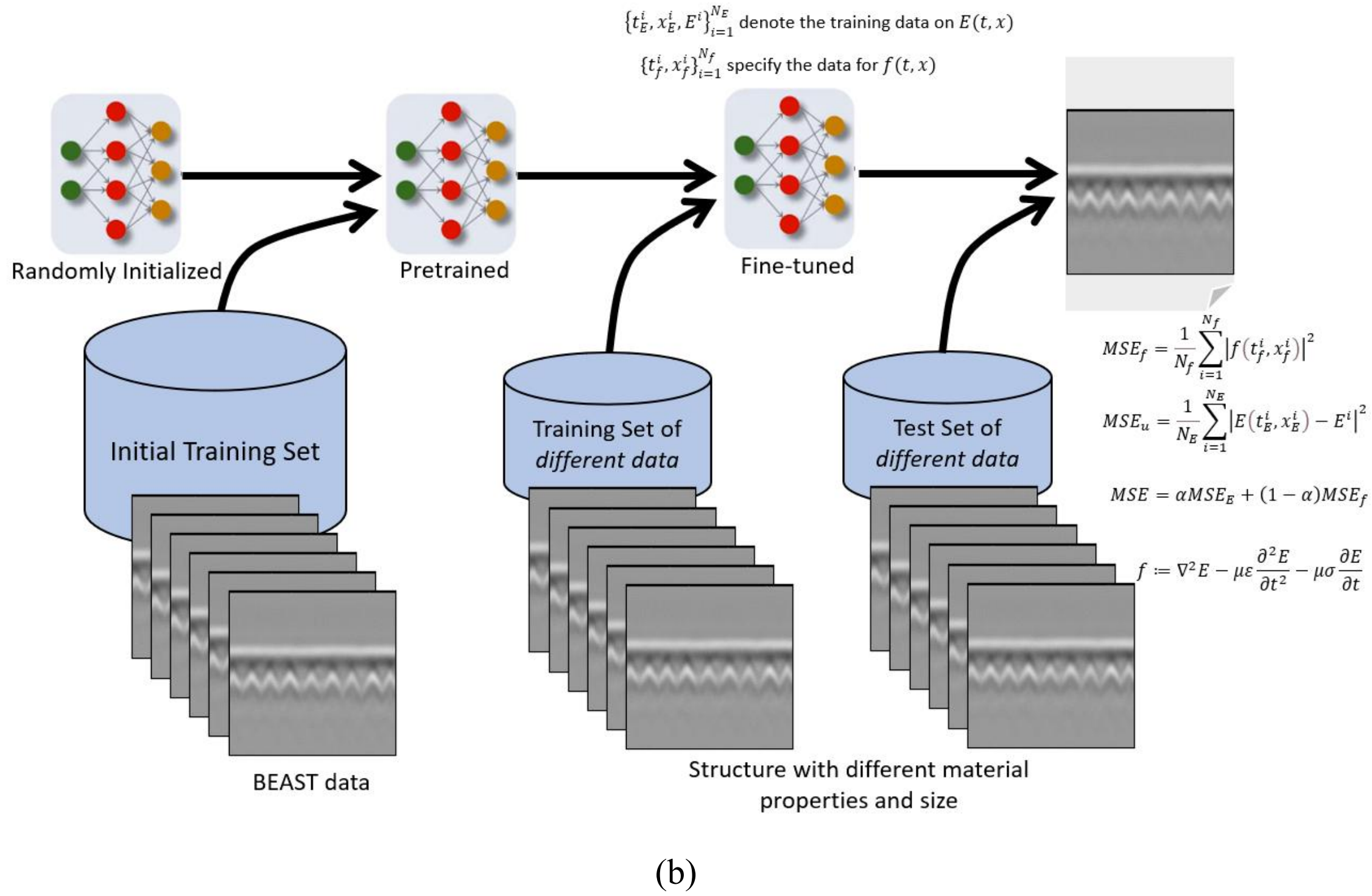

(b)

Figure 12. (a) The schematic of dataset preparation format for model training, (b) The schematic of training process based on initial data and future dataset.

## 6. Performance of PINN

### *6.1. Evaluation Metric and Generalizability*

Generalizability is referred to the ability of a model to perform well on unseen data sets within its training input domain range. The trained network with the lowest validation loss is saved for fine-tuning on the different datasets (prediction/testing). To demonstrate the effectiveness of the different modulus, the comparison of training loss functions is shown in Figure 13. It is seen that the attention blocks improve the validation accuracy. Figure *14* outlines the MSE value of the developed model. The mean value for MSE ranges between 4 and 7, which stands witness to the superior performance of the model. Figure *14* indicates the model's performance over the test dataset. These figures demonstrate that the prediction precisely traces the ground truth. As for the different network configurations, the proposed model has the best prediction results for the *PINN,* with the mean value of MSE being 4. The worst performance is for the CNN, where the model registers a mean MAE value of 70 (Figure *14*).

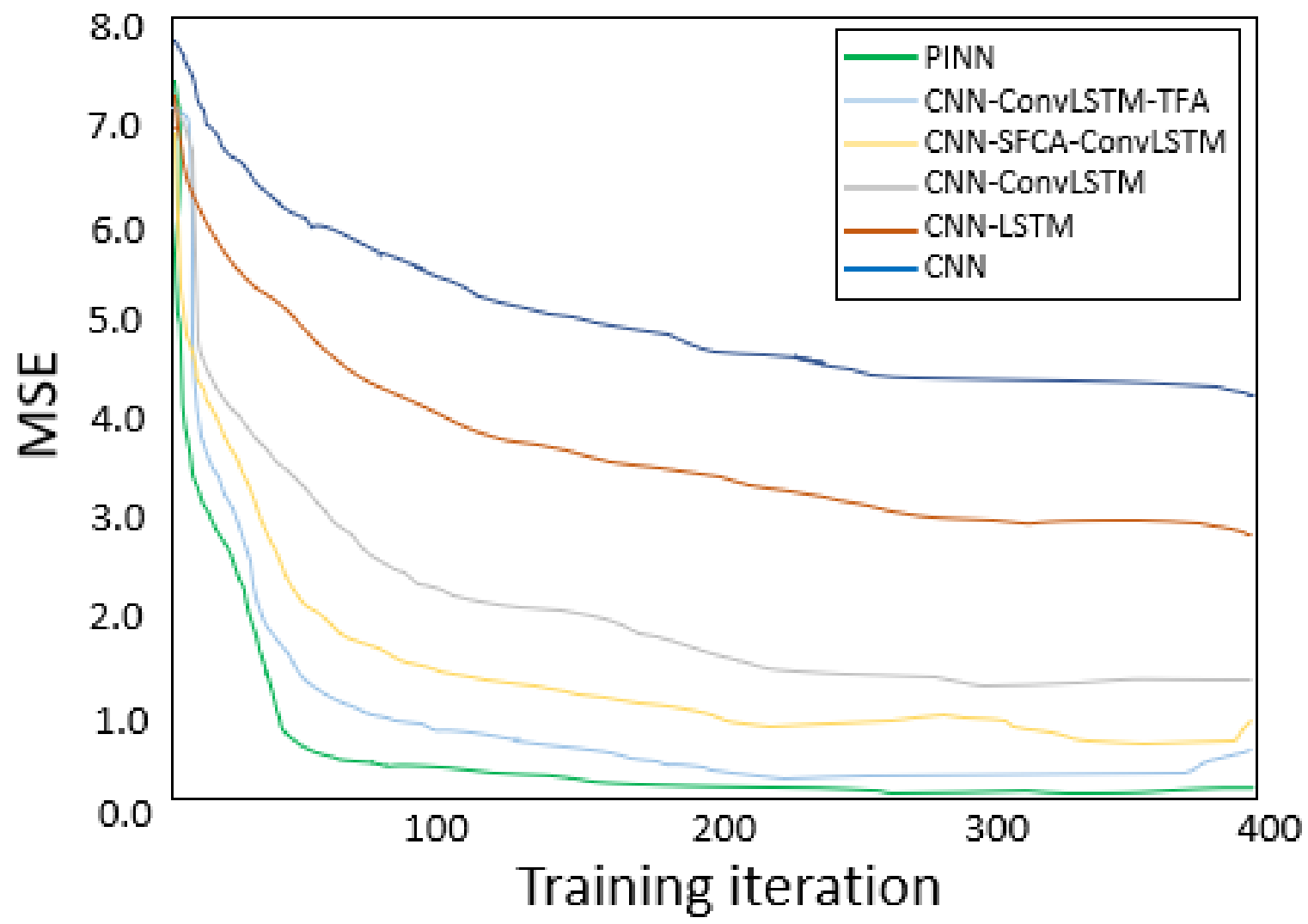

Figure 13. The comparison of training loss functions for different modules in the network

*6.2. Different Modulus to Investigate Performance of PINN*

For verifying the effectiveness of ConvLSTM and attention blocks in the model, the proposed method is compared with the following structures: CNN, CNN-LSTM, CNN-ConvLSTM, CNN-SFCA-ConvLSTM, CNN-ConvLSTM-TFA, and DLVM. The L2 regularization and early-stopping algorithms are applied to account for any overfitting during training. The loss function incorporating the penalties in L2 regularization on layer parameters helps solve overfitting problems during optimization, and L2 defines the regularization term as the sum of the squares of all the feature weights. Early-stopping is another regularization algorithm that can guide how many iterations are required before the model begins to encounter overfitting. The training process is similar to the regression task and MSE is adopted as the loss for the Adam optimizer. The dimension of the input (batch size, time, channel, width, height) is compared to the traditional ConvLSTMs, and in general, they take more time than CNNs to conduct the regressions. The batch size is selected as 4 due to memory exhaustion. To make a fair comparison, the CNN of all methods is based on the proposed CNN. The experimental results are shown in Figure *13* and Figure *14*, in which it can be seen that the *PINN* model achieves the best performance. When CNN-LSTM does not have the attention mechanism, its performance declines substantially, which shows the importance of different areas to different attributes in GPR data analysis recognition and the effectiveness of the attention mechanism. The result of CNN-ConvLSTM shows the effectiveness of ConvLSTM in the extraction of spatial information from related areas. It shows that ConvLSTM has a high activation response to the image area corresponding to the features when predicting the different attributes, indicating that the convolution operation of ConvLSTM affects implicit spatial attention. The spatial channel attention mechanism can re-adjust the weights based on the feature correlation, combined with the spatial correlation ability of ConvLSTM, which results in an effective improvement for the performance of the model. Overall, our proposed method obtained better performance than other methods, which proved the superiority of our method. It is a decently large network with around 14 million parameters, and it requires a good GPU for training, or it ran into memory exhaust errors. The training metrics are presented in Figure 13.

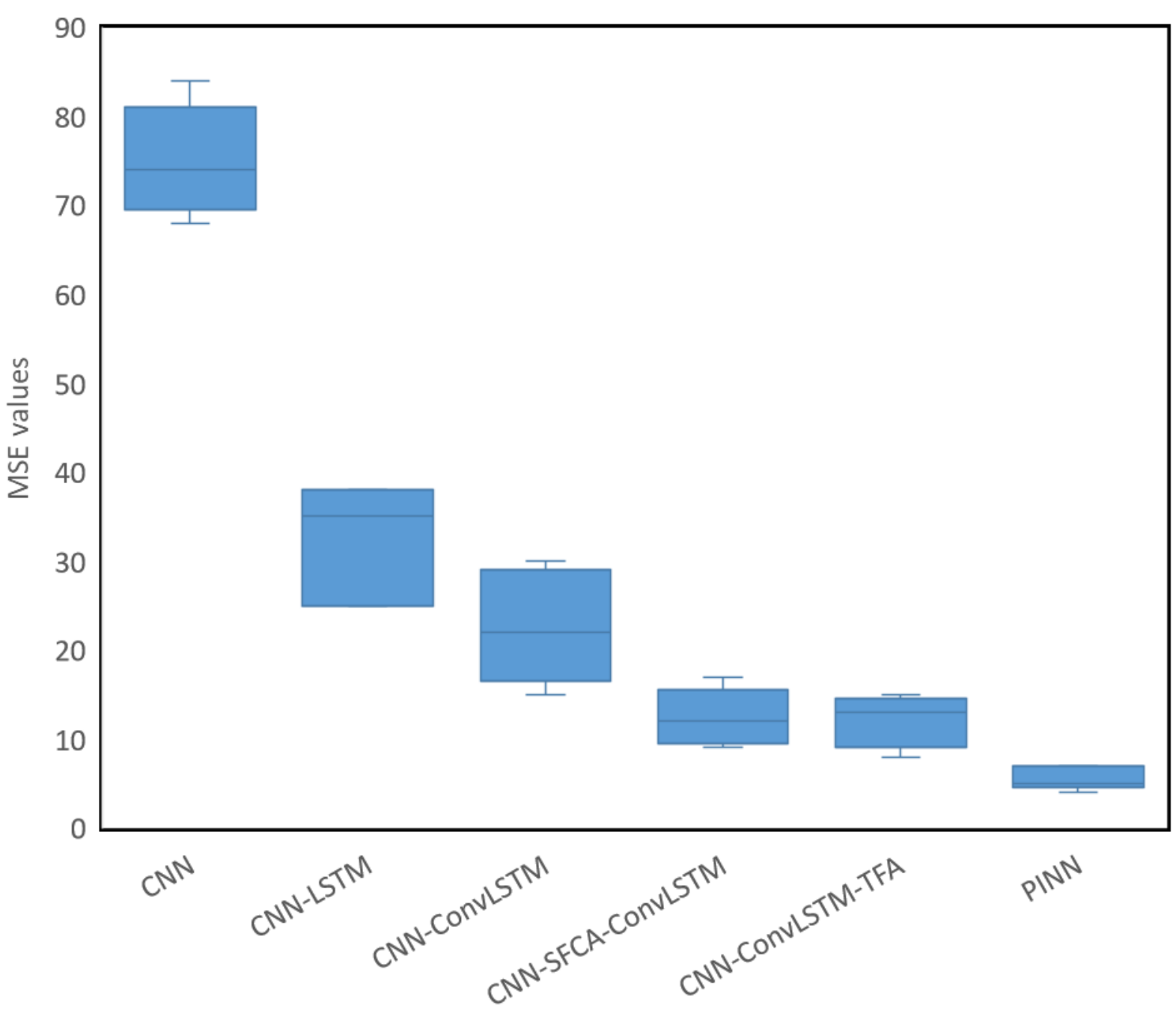

Figure 14. The MSE values, over the validation set.

## 7. Results and discussion

Once the training phase is completed, the developed Physics-Informed Neural Network (PINN) model is prepared to forecast Ground Penetrating Radar (GPR) data using newly introduced datasets to demonstrate its effectiveness and broad applicability. The inputs for this model are samples containing sequences of GPR data formatted as a time series. The model's output is a prediction of the subsequent time step in the GPR data series, which corresponds to the sequence provided as input. This output is generated based on the patterns and dependencies learned by the PINN model during training. This capability to predict future data points showcases the model's ability to not only learn from historical data but also apply this understanding to generate accurate predictions, thereby proving its utility in practical scenarios and its adaptability to different types of GPR data.

The efficacy of the proposed Physics-Informed Neural Network (PINN) model in processing test data is illustrated in Figure 15. The figure highlights the model's ability to forecast future GPR data points by analyzing preceding GPR data sequences. This predictive capacity is showcased through a direct comparison between the actual measured data and the data predicted by the model. Such a comparison not only demonstrates the model's accuracy in predicting future scenarios based on past data but also underlines its potential in practical applications involving GPR data analysis. This ability to accurately forecast future GPR scenarios is critical, as it confirms the model's reliability and enhances its value as a tool in geophysical investigations and other related fields.

To demonstrate how the quantity of training data impacts the model's predictive capabilities, the model undergoes training with various dataset sizes and is subsequently validated using a new test setup. It is evident that as the volume of data used during the training process increases, there is a corresponding improvement in the accuracy of the predictions made by the model. This suggests that incorporating a larger dataset not only enhances the precision of the model's forecasts but also boosts its overall effectiveness. This relationship between data volume and model performance underlines the importance of comprehensive data collection and usage in training to optimize the functionality and reliability of predictive models.

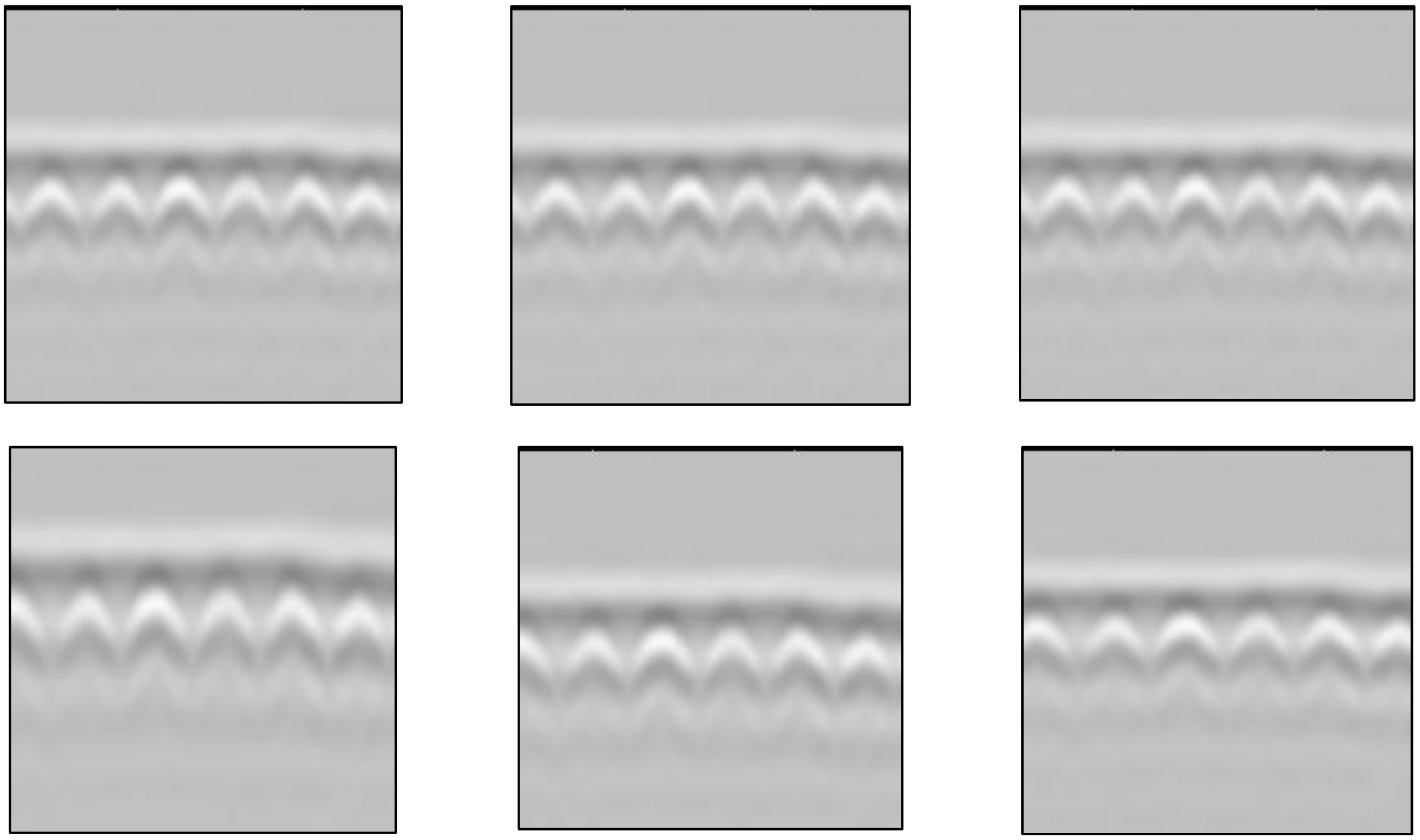

Figure 15. Top row is measurements data and bottom are predicted data using the PINN model, (a) 1000 data samples, (b) 2000 data samples, (c) 3000 data samples.

To illustrate the impact of integrating physics-based principles into the model, the deep learning network is trained without incorporating physics effects, followed by a comparison of the outcomes. The results clearly demonstrate that omitting physics from the model leads to a significant drop in its predictive capabilities. As depicted in Figure 16, this deficiency is particularly noticeable through the loss of information at the edges of the data and a failure to accurately replicate the patterns observed in the actual measurement data. This comparison underscores the critical role that physics plays in enhancing the model's ability to accurately interpret and predict complex data patterns, thereby affirming the value of embedding physical laws into predictive models to ensure more reliable and precise outcomes.

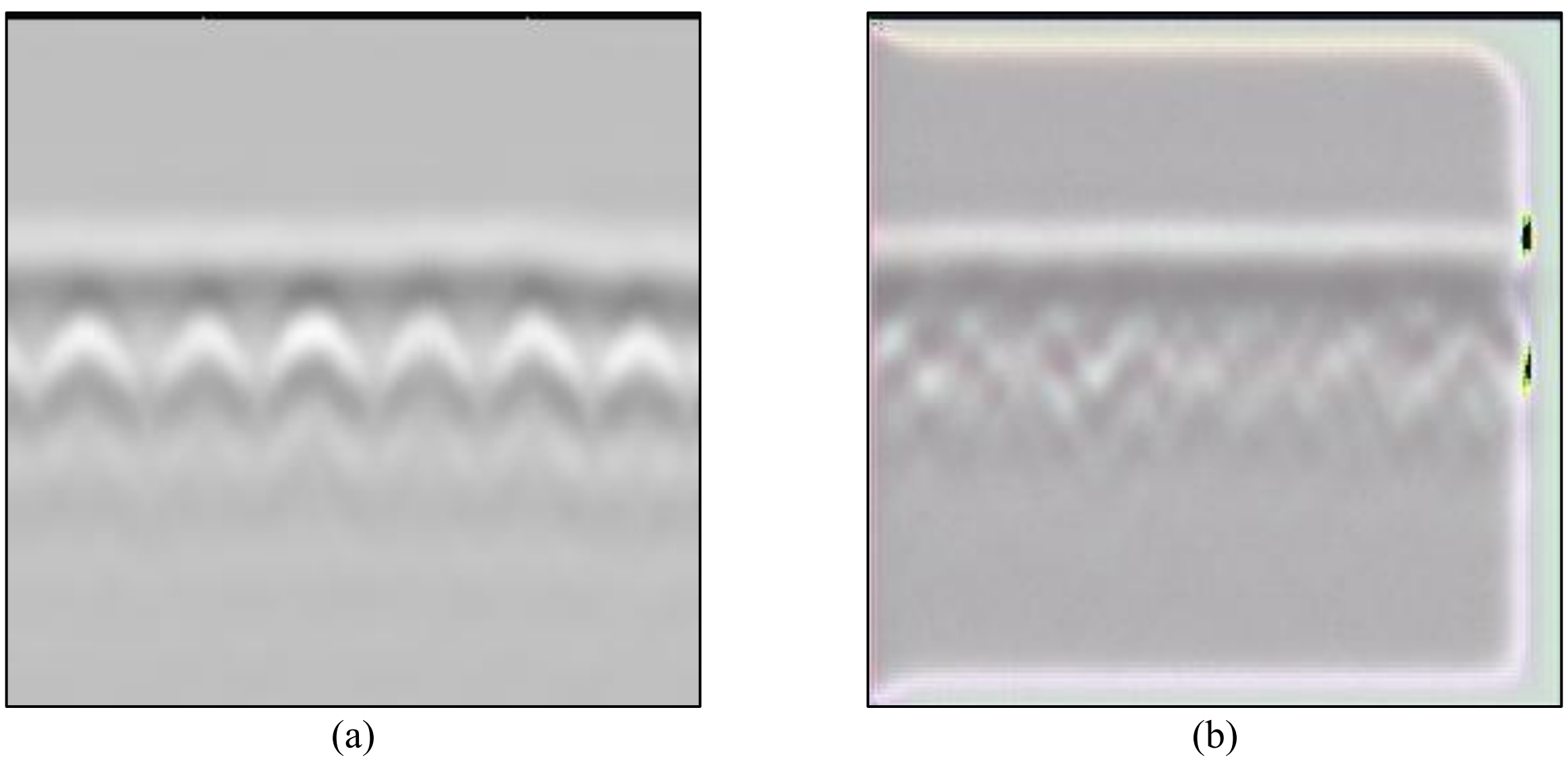
(a) (b)

Figure 16. Comparison of predicted data without physics inclusion

## 8. Ablation Study: Hyperparameter Tuning for DLVM

### *8.1. Different CNN Architecture Study for DLVM*

To create a high-quality model that minimizes a predetermined loss function on a given independent dataset, a learning algorithm's hyperparameters must be tuned. The number of layers, size of the layers, size of the filters, learning rate, and batch size are examples of hyperparameters. However, in this paper, different configurations are selected for the CNNs architectures, including the inception-based CNN, VGG-based architecture, and ResNet-based architecture to measure the performance of the different CNNs for the GPR data prediction task. The MSE values, over the training set and validation set, for three of the selected model configurations are displayed in Figure *17*. All three CNNs hyperparameter configurations are listed in Table *2* and Table *3* and based on the results mentioned in Figure *17*, the most excellent configuration is the inception-based CNN, which is used for the extrapolability and robustness of the viewpoint change test.

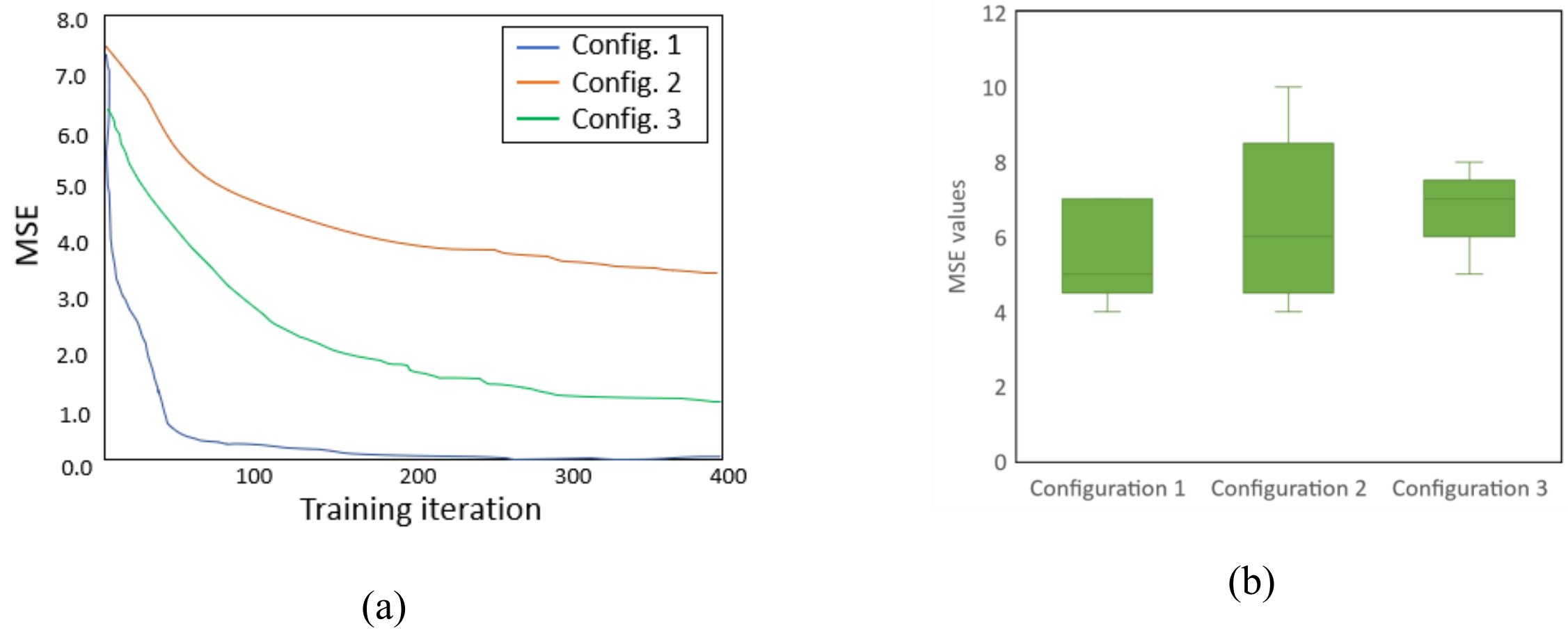

(a) (b)

Figure 17. MAE for training and validation dataset for different configurations (a) the convergence of loss for both training dataset using Adam optimizer, (b) MAE values for validation dataset.

Table 2. Different CNN architecture

| **Configuration** | **Configuration 1** | **Configuration 2** | **Configuration 3** |
|---|---|---|---|
| Convolution layer | K(3,3), Conv(64) | K(3,3), Conv(64) | K(3,3), Conv(64) |
| Max pooling layer | K(3,3) | K(2,2) | K(2,2) |
| Convolution layer | K(3,3), Conv(64) | K(3,3), Conv(128) | K(3,3), Conv(128) |
| Max pooling layer | K(3,3) | K(2,2) | K(2,2) |
| Residual layer | - | One residual layer | - |
| Inception modulus | Inception | - | - |
| Max pooling | K(3,3) | - | |
| Inception modulus | Inception | - | - |
| Convolution layer | - | K(3,3), Conv(128) | K(3,3), Conv(256) |
| Max pooling layer | K(3,3) | K(2,2) | K(2,2) |
| Convolution layer | - | K(3,3), Conv(128) | K(3,3), Conv(512) |
| Max pooling layer | - | K(2,2) | K(2,2) |
| Attention modulus | Channel attention | Channel attention | Channel attention |
| RNN | ConvLSTM | ConvLSTM | ConvLSTM |
| Attention modulus | Temporal attemtion | Temporal attemtion | Temporal attemtion |
| Dense layer | FC (3) | FC (3) | FC (3) |
| Learning rate | 0.001 | 0.001 | 0.001 |
| Batch size | 8 | 8 | 8 |

Table 3. Hyperparameters used for the present *PINN*.

| Parameter | Value | Parameter | Value |
|---|---|---|---|
| CNN filter size | 3×3 | Batch size | 4 |
| CNN pooling size | 2×2 | Optimize for network | Adam |
| Number of layers | 6 | Learning rate of Adam | 0.0001 |
| Number of training data | 322 | $\beta_1$ of Adam | 0.9 |
| Time interval of data | 0.001 | $\beta_2$ of Adam | 0.999 |
| Percentage of training data | 75% | Learning rate decay of Adam | 0.0 |
| Number of epochs | 400 | - | - |

Figure *17* shows the loss function over time, and it can also be noted that the network with the inception module performed better than the VGG and ResNet-based CNN by a good margin. The VGG-based network has been oscillating in its losses. On the other hand, the network with the Inception modules performs better and has a stable convergence comparatively. For VGG, the batch size is 4, while for the inception–based, it is 4 (due to memory exhaustion). The metrics would have been much smoother if a higher batch size was used for the inception model, so the Inception-based model is selected for spatial feature extraction of the images.

*8.2. Study of Epsilon and Alpha Variation during training*

These parameters are defined as a trainable variable, which means it can be adjusted during the training process to better fit the data. This approach improves the model's accuracy by allowing it to select and refine material properties to match real-world data and conditions more closely. The results of varying epsilon and alpha are depicted in Figure 18. Different initial values are chosen for these parameters to avoid introducing bias. As shown in Figure 18, after training, the epsilon value converges to 1, and the alpha value falls within the range of 0.5-0.6. These results suggest that epsilon converges to a specific value tailored to the trained dataset, while alpha converges to a range specific to this dataset.

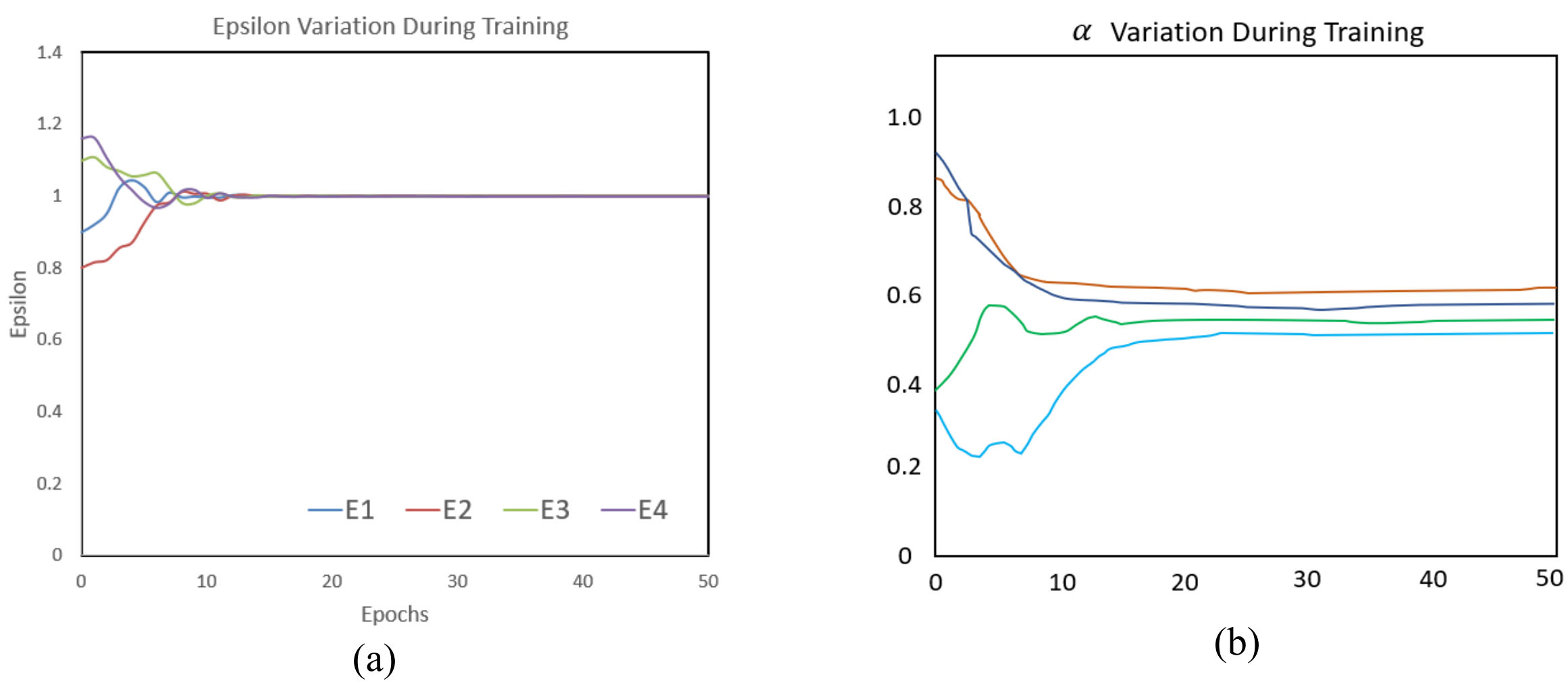

Figure 18. Study of Epsilon and Alpha Variation during training (a) Epsilon variation during training, (b) $\alpha$ variation during training

*8.3. Sequence Size Effect Selection for DLVM*

Input sequence size is a critical hyperparameter that affects the performance of the proposed model. The larger sequence size is more memory efficient for tracking spatio-temporal features and less time-consuming. However, it will memorize the existing features, which in turn decreases its generalization. On the other hand, the smaller sequence size has more generalization capability by tracking more spatio-temporal features of the images at the expense of losing global information and imbibing more comprehensive information. To optimize the sequence size, different sequence sizes are selected, summarized in Table 4, and based on the results shown in Table *4* and Figure 19, the sequence size is selected to be 6 in this paper. Since the whole number of the data is limited, changing the number of the sequence size can affect the whole number of training samples. As it can be seen in Figure 19, by increasing the number of sequences the value of the training error is changes at the expense of losing more data to train the model. The optimal value of the number of sequences is selected to be 6.

Table 4. Sensitivity analysis on different Sequence Size Configurations.

| Sequence Size | Sequence Size | Training | Training |
|---|---|---|---|

| Configuration (SSC#) | (SC) | Error | Time (hr) |
|---|---|---|---|
| 1 | 10 | 0.22 | 3 |
| 2 | 9 | 0.17 | 7 |
| 3 | 8 | 0.12 | 13 |
| 4 | 7 | 0.10 | 22 |
| 5 | 6 | 0.08 | 41 |
| 6 | 5 | 0.13 | 64 |
| 7 | 4 | 0.11 | 71 |
| 8 | 3 | 0.12 | 79 |

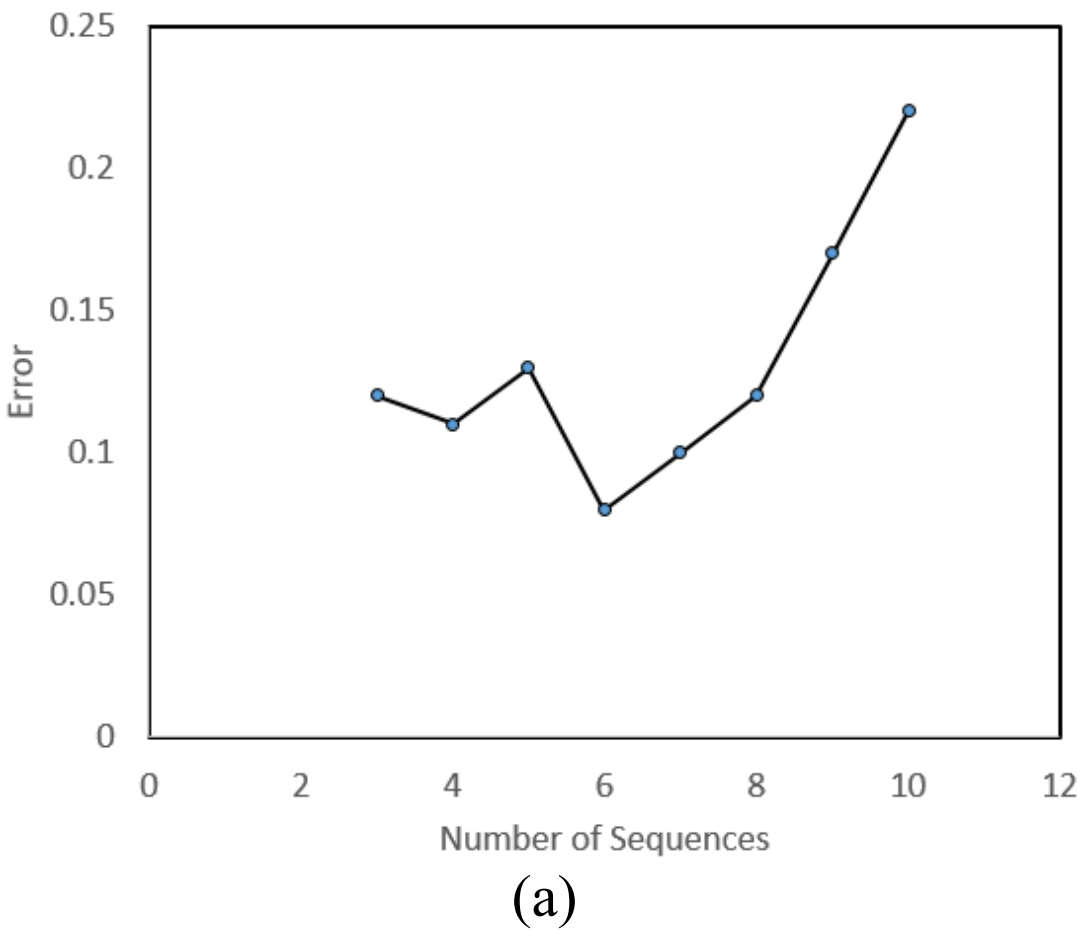

(a)

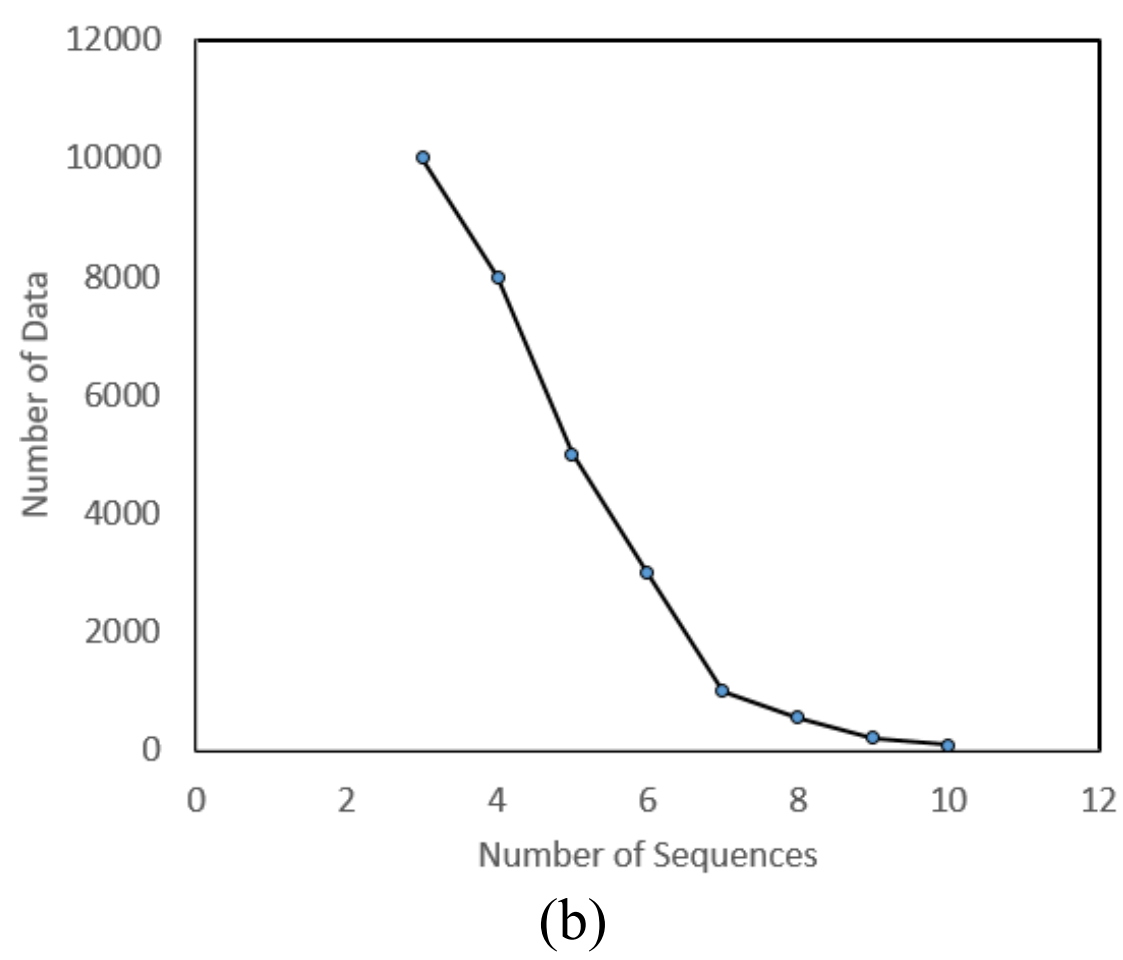

(b)

Figure 19. Sensitivity analysis on different Sequence Size Configuration, (a) MSE values for different SSC configurations; (b) Amount of data versus number of sequences

*8.4. Different Optimizers used to Train DLVM*

The experiment is independently performed for three different optimizers — stochastic gradient descent (SGD), stochastic gradient descent with learning rate decay (SGDLR), and Adam optimizer. The initial learning rate is set to 0.001 for all three optimizers. Results from SGD and Adam are compared in Figure 20 for validation loss and validation accuracy, respectively. The SGDLR produced results very similar to SGD, which are depicted in Figure *20* for clarity. It can be observed that the Adam optimizer converges much more quickly than the SGD optimizer. Hence, the Adam optimizer is selected for all further tests and experiments. A graph showing the convergence of loss function with steps is shown in Figure *20*, respectively.

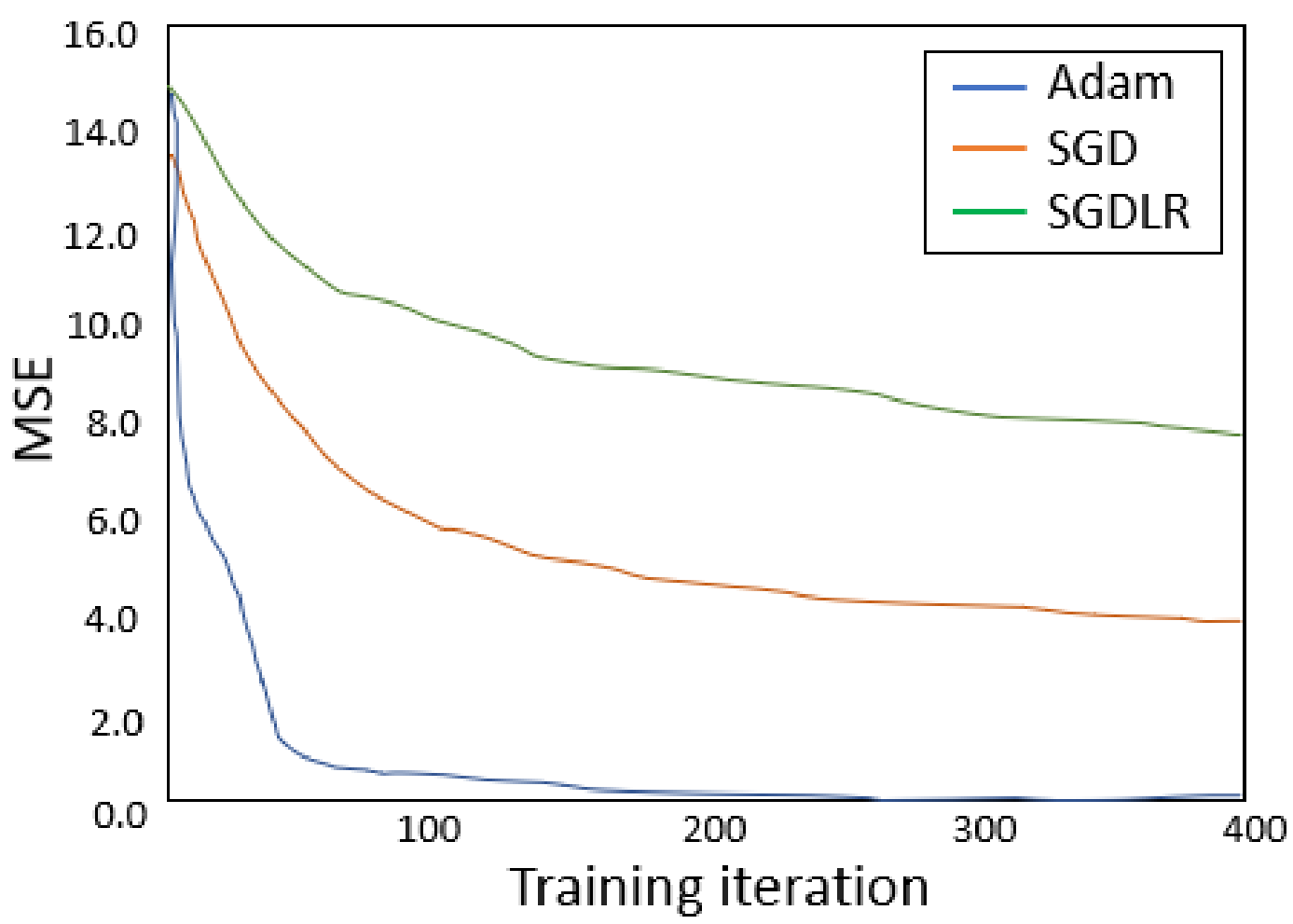

Figure 20. Show the comparison of three optimizers.

**Conclusion**

This research has embarked on advancing the state-of-the-art in Non-Destructive Evaluation (NDE) through the application of Physics-Informed Neural Networks (PINNs), focusing on predicting Ground-Penetrating Radar (GPR) data. Our investigation underscores a significant shift from traditional data-driven models to more sophisticated machine learning frameworks that integrate physical laws directly into their architecture, thus ensuring that predictions not only align with empirical data but also adhere to established physical theories.

Key insights from our research include:

1. **Enhanced Prediction Accuracy**: By incorporating physics directly into the neural network, our model has demonstrated improved accuracy in predicting GPR data, which is crucial for effective bridge deck condition assessments and other civil infrastructure evaluations.
2. **Robust Data Utilization**: The model efficiently leverages existing NDE data, integrating spatial and temporal information, which significantly enhances the depth and reliability of the analysis compared to traditional methods that treat data points in isolation.
3. **Advancements in NDE Technology**: The application of PINNs has shown potential in revolutionizing the field of NDE by enabling more accurate forecasts of infrastructure deterioration and providing a better understanding of underlying deterioration mechanisms through the lens of physics-based models.

**Implications for Future Research and Practice:**

- **Broad Applicability**: The methodology and findings suggest that PINNs can be adapted to a wide range of applications beyond GPR data analysis, including other forms of NDE and different areas of civil engineering.
- **Reduction in Data Needs**: One of the significant advantages of physics-informed models is their ability to achieve high accuracy with less data compared to traditional deep learning models, which require vast datasets to train effectively. This characteristic is particularly beneficial in fields where data collection is expensive or logistically challenging.

- **Tool for Real-Time Analysis**: As these models can swiftly process and predict based on real-time data, they offer potential for deployment in live monitoring systems, providing ongoing assessment without the need for extensive manual analysis.

**Future Directions:**

- **Model Optimization and Validation**: Future work will focus on further optimizing the network architecture and training processes to enhance efficiency and accuracy. Additionally, extensive validation across more diverse datasets will be necessary to generalize the model's applicability.
- **Integration into Existing Systems**: Efforts will be directed towards integrating these models into existing NDE tools and systems, potentially offering a more seamless and automated analysis process for current practitioners.
- **Expanding the Model's Scope**: Exploring the application of PINNs in other related fields, such as seismic analysis and material science, could provide new insights and advancements in those areas.

In conclusion, this research provides a foundational step towards more sophisticated, accurate, and reliable NDE techniques that can significantly impact the maintenance and safety of civil infrastructure. The integration of physics into deep learning presents a promising avenue for future exploration and development in the field of structural health monitoring and beyond.